\documentclass[conference]{IEEEtran}
\IEEEoverridecommandlockouts

\usepackage{cite}
\usepackage{amsmath,amssymb,amsfonts}
\usepackage{algorithmic}
\usepackage{graphicx}
\usepackage{textcomp}
\usepackage{xcolor}
\usepackage{bm}
\usepackage{mathtools, cuted}
\usepackage{comment}

\def\BibTeX{{\rm B\kern-.05em{\sc i\kern-.025em b}\kern-.08em
    T\kern-.1667em\lower.7ex\hbox{E}\kern-.125emX}}

\usepackage{url}
\usepackage{amsthm}

\newtheorem{Proposition}{Proposition}

\newtheorem{Lemma}{Lemma}

\newtheorem{Remark}{Remark}
\newtheorem{Example}{Example}
\newcommand{\argmax}{\mathop{\rm arg~max}\limits}

\usepackage[dvipsnames]{xcolor}
\usepackage{hyperref}
\hypersetup{colorlinks,linkcolor={BrickRed},citecolor={blue},urlcolor={blue}}

\begin{document}

\title{Soft Bayesian Context Tree Models \\ for Real-Valued Time Series
}

\author{
\IEEEauthorblockN{Shota Saito}
\IEEEauthorblockA{
\textit{Gunma University}\\
Gunma, Japan \\
shota.s@gunma-u.ac.jp}
\and
\IEEEauthorblockN{Yuta Nakahara}
\IEEEauthorblockA{
\textit{Waseda University}\\
Tokyo, Japan \\
y.nakahara@waseda.jp}
\and
\IEEEauthorblockN{Toshiyasu Matsushima}
\IEEEauthorblockA{
\textit{Waseda University}\\
Tokyo, Japan \\
toshimat@waseda.jp}
}

\maketitle

\begin{abstract}
This paper proposes the soft Bayesian context tree model (Soft-BCT), which is a novel BCT model for real-valued time series. The Soft-BCT considers soft (probabilistic) splits of the context space, instead of hard (deterministic) splits of the context space as in the previous BCT for real-valued time series. A learning algorithm of the Soft-BCT is proposed based on the variational inference. The results of experiments demonstrate the superiority of the Soft-BCT compared to the previous BCT for some datasets.
\end{abstract}


\section{Introduction} \label{section_introduction}

In the ISIT'93 paper \cite{CTW93} and later in the seminal paper \cite{CTW95}, Willems et al. proposed the context tree weighting (CTW) method. Since the work of Willems et al., the CTW method has been extended and studied in various contexts, including the CTW method for an infinite depth context tree \cite{CTW_98}, context tree model estimation based on MDL criteria\cite{CTM,Gao2006}, estimation of the secrecy-rate of physical unclonable functions (PUFs) \cite{CTW_06}, classification for discrete time series \cite{CTWclassification}, and entropy estimation for discrete time series\cite{Gao2006}, \cite{CTWentropy}, etc.

It has been pointed out that the CTW method can be interpreted from a Bayesian viewpoint. In order to aid the initial discussion, we introduce some definitions. 
These definitions are used only in Section \ref{section_introduction}, and we formally state the definitions in Section \ref{section_model}. Let $\mathcal{T}(D)$ denote the set of all context tree models with maximum depth $D$. Let $T \in \mathcal{T}(D)$ denote a context tree model and $\theta_T$ denote a parameter of $T$ (i.e., $\theta_T$ is a vector consisting of occurrence probabilities of each symbol under a context tree model $T$). From a Bayesian viewpoint, there are two perspectives:
\begin{itemize}
    \item[(i)] $\theta_T$ is assumed to be a random variable, and a prior distribution $p(\theta_T ; T)$ is assumed.
    \item[(ii)] Not only $\theta_T$ but also $T$ is assumed to be a random variable, and prior distributions $p(T)$ and $p(\theta_T | T)$ are assumed.\footnote{To distinguish the perspectives (i) and (ii), we denote $p(\theta_T;T)$ when $T$ is not a random variable and $p(\theta_T | T)$ when $T$ is a random variable. Some papers call the context tree model with $p(T)$ and $p(\theta_T | T)$ ``Bayesian context tree'' (see Remark \ref{remark_BCT}).}
\end{itemize}

In \cite{CTW95}, Willems et al. mentioned that the estimated probability, which is a component of the CTW method and known as the KT-estimator, is obtained by assuming $p(\theta_T;T)$ as a Dirichlet distribution.

In the ISIT'94 paper \cite{CTW_Matsu_94}, Matsushima and Hirasawa considered both $p(T)$ and $p(\theta_T | T)$, and reformulated the CTW method from the viewpoint of the Bayes decision theory. The universal source code designed by the Bayes decision theory is called the Bayes code \cite{TIT_Matsu_91}. There are two types of the Bayes code: one is the non-predictive Bayes code that computes the joint (or block) coding probability $p(x^n)$ given as
\begin{align}
    & p(x^{n}) =  \sum_{T \in \mathcal{T}(D)} p(T)  \int p(\theta_T | T) p(x^{n} | \theta_T, T) \mathrm{d} \theta_T, \label{eq_nonpredictive_coding_prob}
\end{align}
and the other is the predictive Bayes code that computes the predictive coding probability (or the \emph{posterior predictive distribution}) $p(x_{n+1} | x^{n})$ given as
\begin{align}
    & p(x_{n+1} | x^{n}) =  \sum_{T \in \mathcal{T}(D)} p(T | x^{n}) \nonumber \\
    & \quad \times \int p(\theta_T | T, x^{n}) p(x_{n+1} | x^{n}, \theta_T, T) \mathrm{d} \theta_T, \label{eq_predictive_coding_prob}
\end{align}
where $x^{n} = x_1 x_2 \ldots x_{n}$ denotes a source sequence (a discrete-valued time series) of length $n$. The CTW method can be viewed as a special version of the non-predictive Bayes code.

In the ISSPIT'07 paper \cite{CTW_Matsu_07}, Matsushima and Hirasawa proposed a special prior distribution $p(T)$ (see \eqref{eq_prior_T}) and showed that \eqref{eq_predictive_coding_prob} can be computed exactly (i.e., without any approximation) and efficiently (i.e., the computational cost is linear to the length of the sequence). Note that the CTW method computes the joint coding probability $p(x^n)$ and calculates the predictive coding probability by $p(x_{n+1} | x^{n}) = p(x^{n+1})/p(x^{n})$, while Matsushima and Hirasawa \cite{CTW_Matsu_07} directly compute \eqref{eq_predictive_coding_prob}. In the ISIT'09 paper \cite{CTW_Matsu_09}, the algorithm of \cite{CTW_Matsu_07} was extended to an infinite depth context tree model, and the exact and efficient (in terms of both time complexity and space complexity) calculation of \eqref{eq_predictive_coding_prob} was proposed. Moreover, Matsushima et al. applied these ideas to various problems (e.g., \cite{ChangingContextTree, VSBT_arXiv, Quadtree21, Quadtree22, dobashi_entropy21, Nakahara_batch_metatree, nakahara_aistats25, ichijo_mlsp25, rooted_trees}), and performed the theoretical analyses for \eqref{eq_nonpredictive_coding_prob} and \eqref{eq_predictive_coding_prob} in various settings (e.g., \cite{Goto_98, Goto_01, Miya_14, Saito_15, Saito_15_IEICE, Saito_16}). 

Recently, in the ISIT'21 paper \cite{PK21} and later in \cite{kontoyiannis_22}, Kontoyiannis et al. revisited a similar Bayesian interpretation. They proposed the Bayesian context tree (BCT) framework. Kontoyiannis et al. have actively applied the BCT framework to various problems, e.g., maximum posterior probability (MAP) estimation for context tree models\cite{kontoyiannis_22}, change point detection\cite{kontoyiannis_cange_point}, posterior distribution of the BCT \cite{PK_22}, \cite{PK_24}, entropy estimation \cite{PK_23_ITW}, theoretical analyses for the BCT \cite{K_24}, and the BCT model for real-valued time series \cite{PK23}, \cite{PK25}.

Of particular interest to our current study is the BCT model for real-valued time series by Papageorgiou and Kontoyiannis \cite{PK23}, \cite{PK25}. In these previous studies, they used the quantisers from $\mathbb{R}$ to a finite alphabet in order to produce a discrete context. Then, the discrete context follows a deterministic path down the context tree, i.e., the context space divides deterministically. Therefore, the BCT model in \cite{PK23}, \cite{PK25} is restricted by \emph{hard} splits of the context space. In this study, on the other hand, we consider \emph{soft} (probabilistic) splits of the context space. This is a similar idea of a soft decision tree in machine learning (e.g., \cite{HME}, \cite[Chapter 14]{bishop}). As we will describe in Section \ref{section_model}, our soft Bayesian context tree model (Soft-BCT) is a generalized and more flexible model than the BCT-AR model in \cite{PK23}, \cite{PK25}.

We propose a learning method for the Soft-BCT by using variational inference. The results of simulation experiments demonstrate the superiority of the proposed Soft-BCT compared to the previous BCT-AR for some datasets.

\begin{Remark} \label{remark_BCT}
    In \cite{kontoyiannis_22}, the Bayesian context tree is defined as follows (see the first paragraph of Section 1.1 in \cite{kontoyiannis_22}): ``We refer to the models in $\mathcal{T}(D)$ equipped with this prior structure as Bayesian context trees (BCT)''. As described above, the Bayesian context tree in this sense was considered in Matsushima and Hirasawa \cite{CTW_Matsu_94}, \cite{CTW_Matsu_07}, \cite{CTW_Matsu_09}, although they did not use the terminology ``Bayesian context tree.''
\end{Remark}

\section{Soft Bayesian Context Tree Model} \label{section_model}

\begin{figure}[t]
    \centering
    \includegraphics[width=\linewidth]{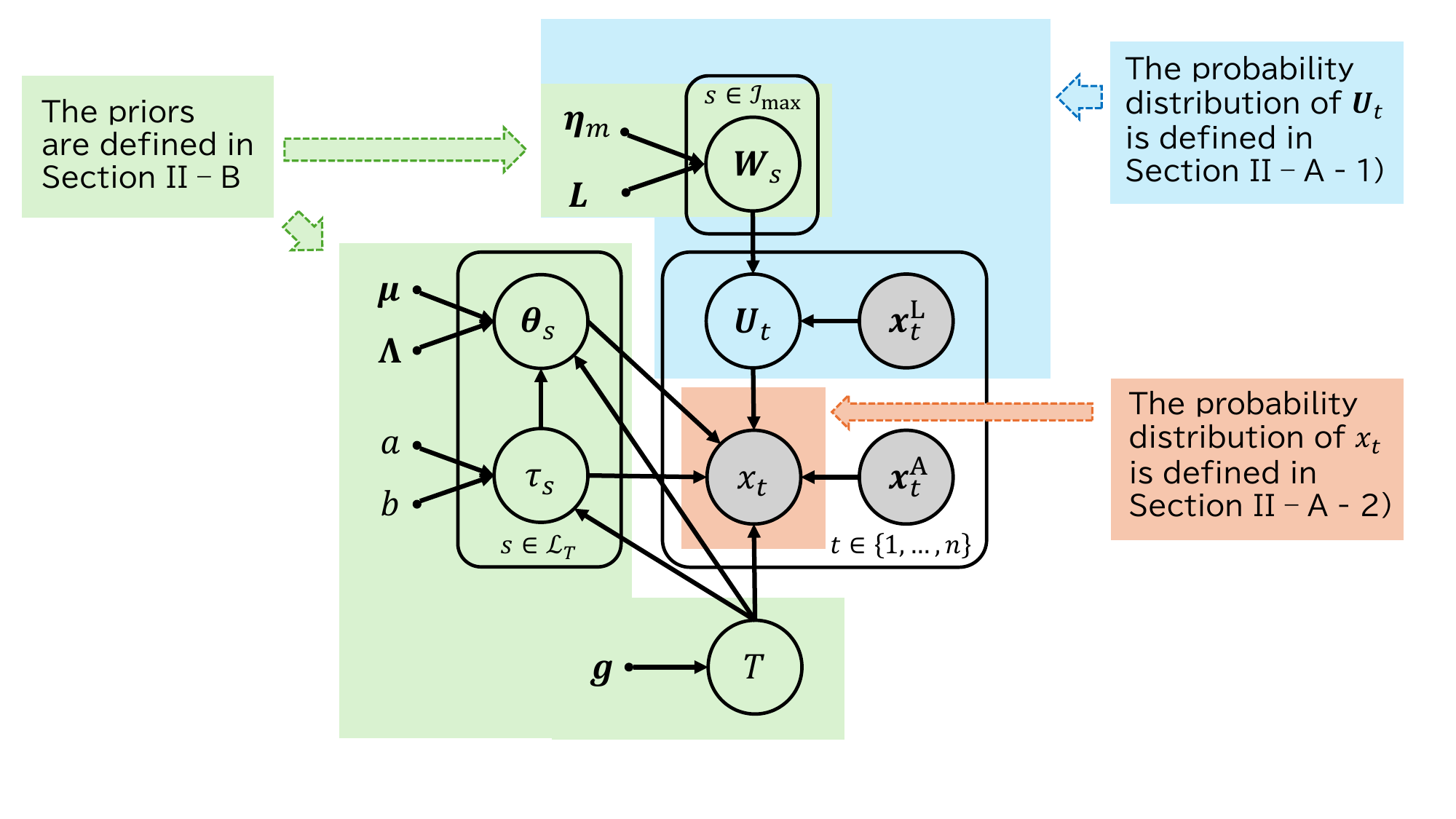}
    \caption{The graphical model of our proposed model. We denote observed variables by shading the corresponding nodes.}
    \label{fig:graphical_model}
\end{figure}

This section explains our proposed model: the \emph{soft Bayesian context tree model} (\emph{Soft-BCT}).  Specifically, Section \ref{subsection_data_generative_model} describes a data generative model and Section \ref{subsection_priors} explains prior distributions. The graphical model in Fig.\ \ref{fig:graphical_model} will be useful for understanding our model.

\subsection{Data generative model} \label{subsection_data_generative_model}

\subsubsection{Soft splits of context space}
Let $T_\mathrm{max}$ be an $M$-ary perfect rooted tree with the root node $s_\lambda$ and the depth $D_\mathrm{max}$, where $M \in \mathbb{N}$ and $D_\mathrm{max} \in \mathbb{N}$ are known constants. For $T_\mathrm{max}$, let $\mathcal{I}_\mathrm{max}$ (resp.\ $\mathcal{L}_\mathrm{max}$) denote a set of all inner nodes (resp.\ leaf nodes) of $T_\mathrm{max}$, and $\mathcal{S}_\mathrm{max} \coloneqq \mathcal{I}_\mathrm{max} \cup \mathcal{L}_\mathrm{max}$. For a node $s \in \mathcal{S}_\mathrm{max}$, $d_s \in \{ 0, 1, \dots , D_\mathrm{max} \}$ denotes the depth of $s$. For $m \in \{1,2, \ldots, M \}$, $s_m$ denotes an $m$-th child node of $s$. The notation $s' \preceq s$ means that $s'$ is an ancestor node of $s$ or $s'$ equals $s$. If we use the notation $s' \prec s$, we do not include the case where $s'$ equals $s$. $\mathrm{Ch}(s)$ denotes the set of all child nodes of $s$.

\begin{figure} [t]
\centering
\includegraphics[width=1.0\linewidth]{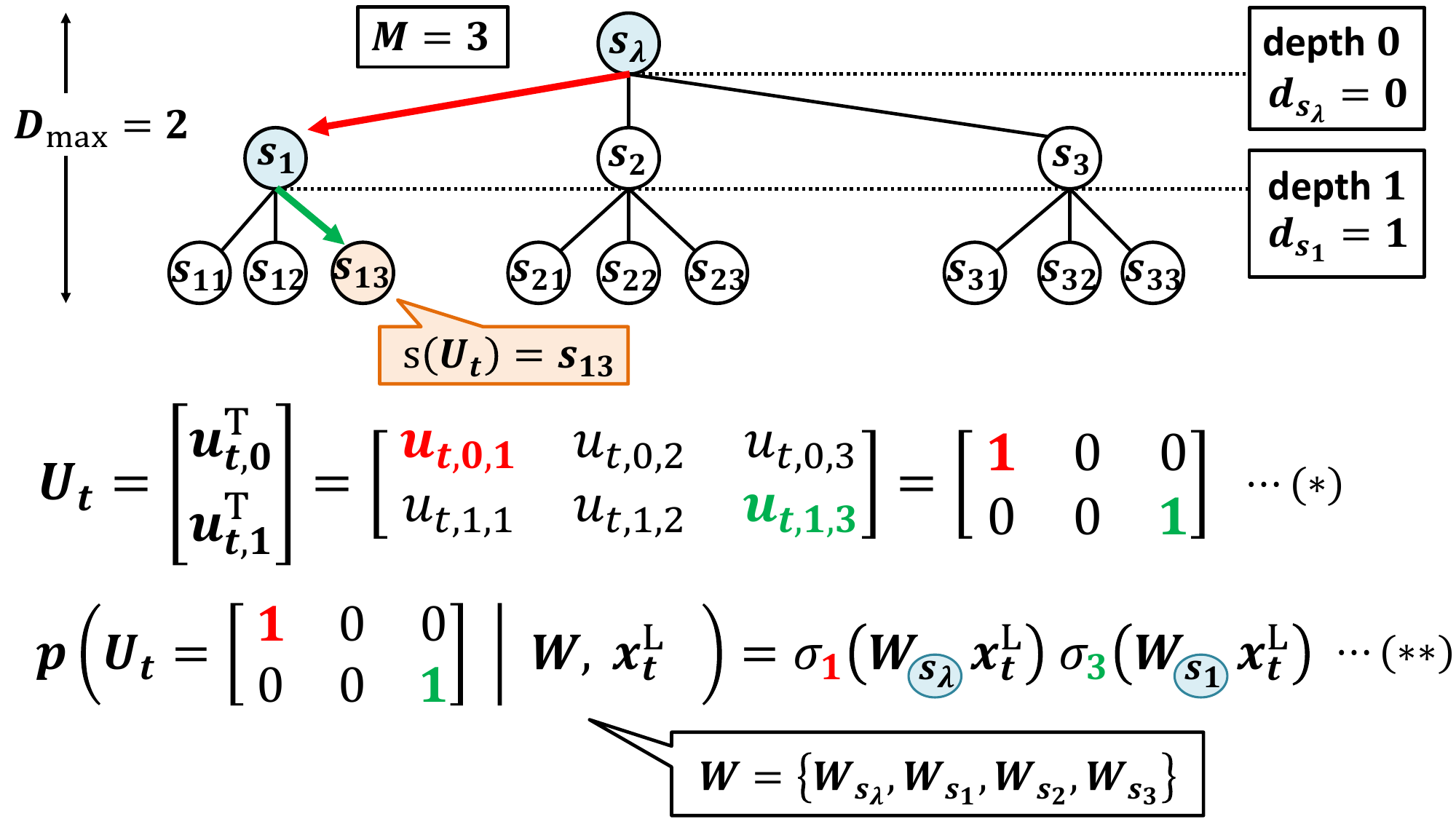}
\caption{An example of $\bm U_t$.}
\label{fig:example}
\end{figure}

We define a matrix indicating a context at time $t \in \{ 1, 2, \dots , n \}$ by 
\begin{align*}
    \bm U_t = [\bm u_{t,0}, \bm u_{t,1}, \dots , \bm u_{t,D_\mathrm{max}-1}]^\top \in \{ 0, 1 \}^{D_\mathrm{max} \times M},
\end{align*}
where $\bm u^\top_{t,d}=[u_{t, d, 1}, u_{t, d, 2}, \ldots, u_{t, d, M}]$ is an $M$-dimensional one-hot vector (i.e., one of the elements equals 1 and all remaining elements equal 0), and $u_{t, d, m}=1$ indicates a path to an $m$-th child node at the branch of depth $d$ (see Fig.\ \ref{fig:example}).

The probability distribution of $\bm U_t$ is defined as
\begin{align}
    p(\bm U_t | \bm W, \bm x_t^\mathrm{L}) = \prod_{s \in \mathcal{I}_\mathrm{max}} \prod_{m=1}^M \sigma_m (\bm W_s \bm x_t^\mathrm{L})^{I \{ s_m \preceq {\sf s}(\bm U_t) \}}, \label{eq_context_model}
\end{align}
where 
\begin{itemize}
    \item $\bm x_t^\mathrm{L} \coloneqq [1, x_{t-1}, x_{t-2}, \cdots, x_{t-J}]^\top \in \mathbb{R}^{J+1}$, where $J \in \mathbb{N}$ is a known constant and ``$\mathrm{L}$'' of $\bm x_t^\mathrm{L}$ stands for logistic regression;
    \item $\bm W_s \in \mathbb{R}^{M \times (J+1)}$ is a coefficient of a logistic regression model, and $\bm W \coloneqq \{ \bm W_s \}_{s \in \mathcal{I}_\mathrm{max}}$. We assume a prior distribution $p(\bm W)$ (see \eqref{eq_prior_W});
    \item  $\sigma_m (\cdot)$ is an $m$-th component of the softmax function, i.e., for an $M$-dimensional vector $\bm v = [v_1, v_2, \ldots, v_M]^\top \in \mathbb{R}^M$, $\sigma_m (\bm v) \coloneqq \exp \{v_m \}/ \sum_{i=1}^{M} \exp \{v_i \}$;
    \item $I \{ \cdot \}$ is the indicator function;
    \item ${\sf s}(\bm U_t)$ is the leaf node of $T_\mathrm{max}$ determined by $\bm U_t$. Note that ${\sf s}(\bm U_t) \in \mathcal{L}_\mathrm{max}$.
\end{itemize}

Please refer to the part shaded in blue in Fig.\ \ref{fig:graphical_model}.
We illustrate \eqref{eq_context_model} in the following example.
\begin{Example}
    As shown in Fig.\ \ref{fig:example}, let $T_\mathrm{max}$ be the perfect rooted tree with $D_\mathrm{max}=2$ and $M=3$. Consider the path in red and green. Then, the matrix $\bm U_t$ corresponding to this path is given by $(\ast)$ in Fig.\ \ref{fig:example}, and ${\sf s}(\bm U_t)=s_{13}$. Moreover, the probability of this $\bm U_t$ is given by $(\ast \ast)$ in Fig.\ \ref{fig:example}.
\end{Example}

\subsubsection{Data generative model of real-valued time series}

We define $\mathcal{T}$ as the set of all regular\footnote{All the nodes have either exactly $M$ children or no children.} rooted sub-trees of $T_\mathrm{max}$ whose root node is $s_\lambda$. For $T \in \mathcal{T}$, we assume a prior distribution $p(T)$ (see \eqref{eq_prior_T}). The set of leaf (resp.\ inner) nodes of $T$ is denoted by $\mathcal{L}_T$ (resp.\ $\mathcal{I}_T$). Each leaf node $s \in \mathcal{L}_T$ has parameters $\bm \theta_s \in \mathbb{R}^{K+1}$ and $\tau_s \in \mathbb{R}_{>0}$, where $K \in \mathbb{N}$ is a known constant, and we denote a collection of these parameters by $\bm \theta \coloneqq \{ \bm \theta_s \}_{s \in \mathcal{L}_T}$ and $\bm \tau \coloneqq \{ \tau_s \}_{s \in \mathcal{L}_T}$. The details of these parameters will be explained shortly. 

Given $x^{t-1}=x_1 x_2 \ldots x_{t-1}$, the probability distribution of $x_t$ is defined as
\begin{align}
    & p(x_t | x^{t-1}, \bm U_t, \bm \theta, \bm \tau, T) \nonumber \\
    & \quad = \prod_{s \in \mathcal{L}_T} \mathcal{N} \left( x_t \Big| (\bm x_t^\mathrm{A})^\top \bm \theta_s, \tau_s^{-1} \right)^{I \{ s \preceq {\sf s}(\bm U_t) \}}, \label{eq_data_generative}
\end{align}
where 
\begin{itemize}
    \item $\mathcal{N}(\cdot | \mu, \sigma^2)$ denotes a probability density function of Gaussian distribution with mean $\mu$ and variance $\sigma^2$;
    \item $\bm x_t^\mathrm{A} \coloneqq [1, x_{t-1}, x_{t-2}, \dots , x_{t-K}]^\top \in \mathbb{R}^{K+1}$, where ``$\mathrm{A}$'' of $\bm x_t^\mathrm{A}$ stands for autoregressive model;
    \item $\bm \theta_s \in \mathbb{R}^{K+1}$ is a coefficient and $\tau_s \in \mathbb{R}_{>0}$ is a precision (inverse of variance) of the autoregressive model. We assume prior distributions of these parameters (see \eqref{eq_prior_theta_tau}).
\end{itemize}

Please refer to the part shaded in red in Fig.\ \ref{fig:graphical_model}.

\subsection{Prior distributions} \label{subsection_priors}
Regarding prior distributions, please refer to the part shaded in green in Fig.\ \ref{fig:graphical_model}.

Let $\bm w_{s,m}^\top$ be a $(J+1)$-dimensional row vector corresponding to the $m$-th row of $\bm W_s$. Then, the prior distribution of $\bm W = \{ \bm W_s \}_{s \in \mathcal{I}_\mathrm{max}}$ is defined as
\begin{align}
    p(\bm W) = \prod_{s \in \mathcal{I}_\mathrm{max}} \prod_{m=1}^M \mathcal{N}(\bm w_{s,m} | \bm \eta_m, \bm L^{-1}), \label{eq_prior_W}
\end{align}
where $\bm \eta_m \in \mathbb{R}^{J+1}$ is the mean and $\bm L \in \mathbb{R}^{(J+1) \times (J+1)}$ is the symmetric positive definite precision matrix .

The prior distribution of $T \in \mathcal{T}$ is defined as
\begin{align}
    p(T) = \left( \prod_{s \in \mathcal{I}_T} g_s \right) \left( \prod_{s \in \mathcal{L}_T} (1-g_s) \right), \label{eq_prior_T}
\end{align}
where $g_s \in [0,1]$ is a hyperparameter on $s \in \mathcal{S}_\mathrm{max}$ and we assume that $g_s=0$ for $s \in \mathcal{L}_\mathrm{max}$. 

\begin{Remark}
    The prior in \eqref{eq_prior_T} was introduced by \cite{CTW_Matsu_07}, \cite{CTW_Matsu_09}, and the properties of \eqref{eq_prior_T}, e.g., $\sum_{T \in \in \mathcal{T}} p(T) = 1$, were summarized in \cite{full_rooted_trees}. After these papers, \eqref{eq_prior_T} was revisited by \cite{PK_22}, \cite{PK_24}. In \cite{PK23}, \cite{PK25}, a similar prior as in \eqref{eq_prior_T} was used, and the parameter $\beta$ in \cite{PK23}, \cite{PK25} corresponds to $1-g_s$ in \eqref{eq_prior_T}.
\end{Remark}

Given $T \in \mathcal{T}$, the prior distribution of $\bm \theta = \{ \bm \theta_s \}_{s \in \mathcal{L}_T}$ and $\bm \tau = \{ \tau_s \}_{s \in \mathcal{L}_T}$ is defined as
\begin{align}
    p(\bm \theta, \bm \tau | T) = \prod_{s \in \mathcal{L}_T} \mathcal{N}(\bm \theta_s | \bm \mu, (\tau_s \bm \Lambda)^{-1}) \mathrm{Gam}(\tau_s | a, b), \label{eq_prior_theta_tau}
\end{align}
where $\bm \mu \in \mathbb{R}^{K+1}$ is the mean and $\bm \Lambda \in \mathbb{R}^{(K+1) \times (K+1)}$ is the symmetric positive definite precision matrix; $\mathrm{Gam}(\cdot | a, b)$ denotes the probability density function of Gamma distribution with parameters $a, b \in \mathbb{R}_{>0}$.

\begin{Remark} \label{remark_model}
    The Soft-BCT model reduces to the soft version of the BCT-AR model in \cite{PK23}, \cite{PK25} when $D_\mathrm{max}=J$ and $\bm w_{s,m}^\top = [w_{s, m, 1}, 0, \ldots, 0, w_{s, m, d_s+2}, 0, \ldots, 0]$. Also, note that the previous BCT-AR model in \cite{PK23}, \cite{PK25} uses the same thresholds at all nodes, while our model can use \emph{different thresholds} at each node.
\end{Remark}

\section{Variational Inference of Soft-BCT}

\subsection{Problem setup} 

Let $\bm x \coloneqq [x_1, x_2, \ldots, x_n]^\top \in \mathbb{R}^{n}$ denote a real-valued time series data. Also, let $\bm U \coloneqq \{ \bm U_t \}_{t=1}^{n}$. We estimate $\bm U$, $T$, $\bm \theta$, $\bm \tau$, and $\bm W$ from $\bm x$. When we assume the 0-1 loss, the optimal decision based on the statistical decision theory is given by MAP (maximum posterior) estimation for $(\bm U, T, \bm \theta, \bm \tau, \bm W)$ (see, e.g., \cite{berger1985statistical}). Therefore, our goal is to derive a posterior distribution of $(\bm U, T, \bm \theta, \bm \tau, \bm W)$ given $\bm x$.\footnote{Strictly speaking, an initial context $x_{1-\max \{J,K \}}, \ldots, x_0$ is also given together with $\bm x$.}

However, it is difficult to find an analytical solution of the posterior distribution  $p(\bm U, T, \bm \theta, \bm \tau, \bm W | \bm x)$. Hence, we use a technique of variational inference to approximate the posterior distribution.

\begin{Remark}
    When $M=2$, we can calculate an approximate posterior distribution of $(\bm U, T, \bm \theta, \bm \tau, \bm W)$ by using local variational methods \cite{bishop}. When $M>2$, however, it is difficult to derive a posterior distribution of $\bm W$ because an efficient variational inference method is not known for a multiclass logistic regression model. In this study, we use a MAP estimator of $\bm W$. Given $\bm W$, we can derive an approximate posterior distribution of $(\bm U, T, \bm \theta, \bm \tau)$.
\end{Remark}

\subsection{Variational inference} 

Given $\bm W$, we approximate the joint posterior distribution $p( \bm U, T, \bm \theta, \bm \tau | \bm x, \bm W )$ by \emph{variational distribution} $q(\bm U, T, \bm \theta, \bm \tau)$, which satisfies the factorization property: $q(\bm U, T, \bm \theta, \bm \tau) = q(\bm U)q(T, \bm \theta, \bm \tau)$.

It is known (e.g., \cite{bishop}) that minimizing the Kullback-Leibler divergence between $q(\bm U, T, \bm \theta, \bm \tau)$ and $p( \bm U, T, \bm \theta, \bm \tau | \bm x, \bm W )$ is equivalent to maximizing the \emph{variational lower bound} $\mathrm{VL}(q;\bm W)$:
\begin{align*}
    \mathrm{VL}(q;\bm W) \coloneqq \mathbb{E}_{q(\bm U, T, \bm \theta, \bm \tau)} \left[ \ln \frac{p(\bm x, \bm U, T, \bm \theta, \bm \tau | \bm W)}{q(\bm U, T, \bm \theta, \bm \tau)}\right].
\end{align*}
Moreover, it is also known (e.g., \cite{bishop}) that the optimal variational distribution satisfies
\begin{align}
&\ln q^* (\bm U) = \mathbb{E}_{q^* (T, \bm \theta, \bm \tau)} \bigl[\ln p(\bm x, \bm U, T, \bm \theta, \bm \tau | \bm W) \bigr] + \mathrm{const.}, \label{eq_q_star_u} \\
&\ln q^* (T, \bm \theta, \bm \tau) = \mathbb{E}_{q^* (\bm U)} \bigl[\ln p(\bm x, \bm U, T, \bm \theta, \bm \tau | \bm W) \bigr] + \mathrm{const.} \label{eq_q_star_T_theta_tau}
\end{align}

Simultaneously, we calculate
\begin{align}
    \bm W^* \coloneqq \argmax_{\bm W} ~ \left\{ \mathrm{VL}(q; \bm W) + \ln p(\bm W) \right \}. \label{eq_estimator_W}
\end{align}
The estimator given by \eqref{eq_estimator_W} can be seen as an approximate MAP estimator of $\bm W$ (e.g., \cite[Chapters 9 and 10]{bishop}). For Bayesian context tree models, a similar technique as in \eqref{eq_estimator_W} was used in our previous study \cite{NakaharaISIT23}.

However, $q^* (\bm U)$, $q^* (T, \bm \theta, \bm \tau)$, and $\bm W^*$ depend on each other. Therefore, we update $q (\bm U)$, $q (T, \bm \theta, \bm \tau)$, and $\bm W$ in turn from an initial value until convergence. We show the update formula of \eqref{eq_q_star_u} and \eqref{eq_q_star_T_theta_tau} in Section \ref{section_update_parameters} and the update formula of \eqref{eq_estimator_W} in Section \ref{section_update_W}.

\subsection{Update formula of \texorpdfstring{\eqref{eq_q_star_u}}{(8)} and \texorpdfstring{\eqref{eq_q_star_T_theta_tau}}{((9))}} \label{section_update_parameters}

The update formula of $q(\bm U)$ is given by Proposition \ref{prop_q(U)}, which follows from Lemma \ref{lem_q_U} in Appendix \ref{section_appendix_posterior} and Lemma \ref{lem_expectations} in Appendix \ref{section_appendix_expectation}.

\begin{Proposition} \label{prop_q(U)}
The posterior $q(\bm U)$ can be factorized as $q(\bm U) = \prod_{t=1}^{n} q(\bm U_t)$, and each $q(\bm U_t)$ is 
    \begin{align*}
        q(\bm U_t) = \prod_{s \in \mathcal{I}_\mathrm{max}} \prod_{m=1}^M (\pi'_{t,s,s_m})^{I \{ s_m \preceq {\sf s}(\bm U_t) \}}, 
    \end{align*}
where 
\begin{align*}
    \pi'_{t, s, s_m} \coloneqq \frac{\rho_{t, s, s_m}}{\sum_{m=1}^M \rho_{t, s, s_m}},
\end{align*}
and $\rho_{t, s, s_m} $ is defined as 
\begin{align*}
    &\ln \rho_{t, s, s_m} \coloneqq \begin{cases}
        \ln \sigma_m (\bm W_s \bm x_t^\mathrm{L}) + (\star) \\ 
        \quad + \ln \sum_{s_\mathrm{ch} \in \mathrm{Ch}(s_m)} \rho_{t,s_m,s_\mathrm{ch}}, & s_m \in \mathcal{I}_\mathrm{max}, \\
            \ln \sigma_m (\bm W_s \bm x_t^\mathrm{L}) + (\star), & s_m \in \mathcal{L}_\mathrm{max},
        \end{cases} 
\end{align*}
where $(\star)$ is given as
\begin{align}
    (\star) = \frac{1}{2} & (1-g'_{s_m}) \left(\prod_{\tilde{s} \prec s_m} g'_{\tilde{s}} \right) \Big \{ (-\ln 2\pi + \psi(a'_{s_m}) - \ln b'_{s_m}) \nonumber \\
    & - \frac{a'_{s_m}}{b'_{s_m}}(x_t - (\bm x_t^\mathrm{A})^\top \bm \mu'_{s_m})^2 - (\bm x_t^\mathrm{A})^\top (\bm \Lambda'_{s_m})^{-1} (\bm x_t^\mathrm{A}) \Big \}. \label{eq_def_star}
\end{align}
Here, $\psi(\cdot)$ denotes the digamma function and $\bm \Lambda'_{s_m}$, $\bm \mu'_{s_m}$, $a'_{s_m}$, $b'_{s_m}$, $g'_{s_m}$ are given as in \eqref{Lambda}, \eqref{mu}, \eqref{a}, \eqref{b}, \eqref{g}, respectively.
\end{Proposition}

Next, we show the update formula of \eqref{eq_q_star_T_theta_tau}. To this end, we introduce two notations. Let $\bm Q_s$ be defined as
\begin{align*}
    \bm Q_s \coloneqq \mathrm{diag} \left \{ q_{s,1}, q_{s,2}, \dots , q_{s,n} \right \},
\end{align*}
where $\mathrm{diag} \{ \cdot \}$ denotes a diagonal matrix and $q_{s,t}$ for $t=1,2,\ldots, n$ is defined as 
\begin{align*}
    q_{s,t} \coloneqq \prod_{s', s'_\mathrm{ch} \preceq s} \pi'_{t, s', s'_\mathrm{ch}}. 
\end{align*}
Here, $s'_\mathrm{ch}$ denotes a child node of $s'$, and $q_{s_\lambda,t}=1$. For example, in Fig.\ \ref{fig:example}, $q_{s_{13},t}$ is given as
\begin{align*}
    q_{s_{13},t} = \pi'_{t, s_\lambda, s_1} \pi'_{t, s_1, s_{13}}.
\end{align*}
Let $\bm X^\mathrm{A}$ denote the $n \times (K+1)$ matrix whose $t$-th row is $(\bm x_t^\mathrm{A})^\top$.

The update formula of $q(T, \bm \theta, \tau)$ is given by Proposition \ref{prop_q(T,theta,tau)}, which follows from Lemma \ref{lem_q_T_theta_tau} in Appendix \ref{section_appendix_posterior} and Lemma \ref{lem_expectations} in Appendix \ref{section_appendix_expectation}.

\begin{Proposition} \label{prop_q(T,theta,tau)}
The posterior $q(T, \bm \theta, \tau)$ can be factorized as $q(T, \bm \theta, \bm \tau) = q(T) \prod_{s \in \mathcal{L}_T} q(\bm \theta_s, \tau_s)$. For each $s \in \mathcal{L}_T$, $q(\bm \theta_s, \tau_s)$ is 
\begin{align*}
    q(\bm \theta_s, \tau_s) = \mathcal{N}(\bm \theta_s | \bm \mu'_s, (\tau_s \bm \Lambda'_s)^{-1}) \mathrm{Gam}(\tau_s | a'_s, b'_s),
\end{align*}
where 
\begin{align}
        \bm \Lambda'_s &\coloneqq \bm \Lambda + (\bm X^\mathrm{A})^\top \bm Q_s \bm X^\mathrm{A}, \label{Lambda}\\
        \bm \mu'_s &\coloneqq \left( \bm \Lambda'_s \right)^{-1} \left( \bm \Lambda \bm \mu + (\bm X^\mathrm{A})^\top \bm Q_s \bm x \right), \label{mu}\\
        a'_s &\coloneqq a + \frac{1}{2}\mathrm{Tr} \{ \bm Q_s \}, \label{a}\\
        b'_s &\coloneqq b + \frac{1}{2} \left( \bm \mu^\top \bm \Lambda \bm \mu + \bm x^\top \bm Q_s \bm x - (\bm \mu'_s)^\top \bm \Lambda'_s \bm \mu'_s \right), \label{b}
\end{align}
and $\mathrm{Tr} \{\cdot\}$ denotes the trace of a matrix. Moreover, the update formula of $q(T)$ is given as
\begin{align*}
    q(T) = \left( \prod_{s \in \mathcal{I}_T} g'_s \right) \left( \prod_{s \in \mathcal{L}_T} (1-g'_s) \right), 
\end{align*}
where 
\begin{align}
    g'_s &\coloneqq \begin{cases}
    \frac{g_s \prod_{s_\mathrm{ch} \in \mathrm{Ch}(s)} \phi_{s_\mathrm{ch}}}{\phi_s}, & s \in \mathcal{I}_\mathrm{max}, \\
    0, & s \in \mathcal{L}_\mathrm{max},
    \end{cases} \label{g}\\
    \phi_s &\coloneqq \begin{cases}
    (1-g_s) \gamma_s + g_s \prod_{s_\mathrm{ch} \in \mathrm{Ch}(s)} \phi_{s_\mathrm{ch}}, & s \in \mathcal{I}_\mathrm{max}, \\
    \gamma_s, & s \in \mathcal{L}_\mathrm{max}.
    \end{cases} \label{eq_update_formula_Tree} \\
    \ln \gamma_s &= \frac{1}{2} \ln |\bm \Lambda| - \frac{1}{2} \ln |\bm \Lambda'_s| + a \ln b - a'_s \ln b'_s \nonumber \\
    & \quad - \ln \Gamma (a) + \ln \Gamma (a'_s) - \frac{1}{2} \mathrm{Tr} \{ \bm Q_s \} \ln 2\pi. \label{gamma}
\end{align}
\end{Proposition}

\begin{Remark}
    The formula \eqref{eq_update_formula_Tree} has a similar structure to the CTW algorithm \cite{CTW95} (to be more precise, the Bayes coding algorithm for context tree models \cite{CTW_Matsu_07}, \cite{CTW_Matsu_09}).
\end{Remark}

\subsection{Update formula of \texorpdfstring{\eqref{eq_estimator_W}}{(10)}} \label{section_update_W}

We can use the same algorithm as the learning algorithm of the multiclass logistic regression in \cite[Section 4.3.4]{bishop}. Let
\begin{align*}
    \bm w_s \coloneqq (\bm w_{s,1}^\top, \bm w_{s_2}^\top, \dots , \bm w_{s,M}^\top)^\top \in \mathbb{R}^{(J+1)M},
\end{align*}
where $\bm w_{s,m}$ has been defined in Section \ref{subsection_priors}. Then, the update formula of $\bm w_s$ is given as
\begin{align}
    \bm w_s^\mathrm{(new)} = \bm w_s^\mathrm{(old)} - \bm H_{\bm w_s^\mathrm{(old)}}^{-1} \nabla E(\bm w_s^\mathrm{(old)}), \label{eq_update_formula_w}
\end{align}
where $E(\bm w_s)$ is 
\begin{align}
    E(\bm w_s) = & -\sum_{t=1}^n q_{s,t}  \sum_{m=1}^M \pi'_{t, s, s_m} \ln \sigma_m (\bm W_s \bm x_t^\mathrm{L}) \nonumber \\
    & \quad + \frac{1}{2}\sum_{m=1}^M (\bm w_{s,m} - \bm \eta_m)^\top \bm L (\bm w_{s,m} - \bm \eta_m), \label{eq_def_E}
\end{align}
and $\bm H_{\bm w_s}$ denotes the Hessian matrix of $E(\bm w_s)$. For details, see Appendix \ref{section_appendix_w}.

\section{Experiments}

We compare the Soft-BCT with the previous BCT-AR \cite{PK23}, \cite{PK25}, and numerically confirm the effect of soft-splitting on MAP models and prediction accuracy. 

\subsection{Experiment 1: MAP estimation}

First, we confirm the effect of soft-splitting on the MAP model for the dataset \texttt{unemp}, which is also used in \cite{PK25}. 
We use the following values as the hyperparameters other than $\bm \eta_m$: $D_\mathrm{max} = 10$, $g_s=2^{-m}$ for any $s$, $\bm \mu = \bm 0$, $\bm \Lambda = \bm I$, $a = 0.1$, and $b = 0.1$.\footnotemark[4] 
In this experiment, we assume $J=D_\mathrm{max}$ and use only the $(d_s+2)$-th component of $\bm x_t^\mathrm{L}$ at each internal node $s$ as in Remark \ref{remark_model}. Therefore, $\bm \eta_m$ is represented as a two-dimensional vector $[\eta_{m,1}, \eta_{m,2}]^\top$. For $\bm \eta_m$, we use the following values:
\begin{align*}
    &\bm \eta_m^\top = \nonumber \\
    &\begin{cases}
        [0,0], & m = M\\
        [\eta_{m+1,1} + h_m(\eta_{m+1,2} - \eta_{m,1}), -C(M-m)], & m < M,
    \end{cases}
\end{align*}
where $h_1, \dots, h_{M-1} \in \mathbb{R}$ and $C \in \mathbb{R}_{>0}$ are tuning parameters, which determine the thresholds and steepness of the logistic curve, respectively. In this experiment, we set $h_1 = 0.15$ (the same value used for BCT-AR in \cite{PK25}) and $C=10$. In the variational inference, we used the posterior distribution obtained by using BCT-AR as the initial values of $q(\bm U)q(\bm \theta, \bm \tau, T)$ (i.e., $q(\bm U)$ represents a hard thresholding with probability 1), and we set the initial value of $\bm w_{s,m}$ to $\bm \eta_m$ for each $m$. After convergence, the MAP model is calculated by the MAP tree estimation algorithm in \cite{full_rooted_trees}.

\begin{figure}[h]
    \centering
    \includegraphics[width=0.8\linewidth]{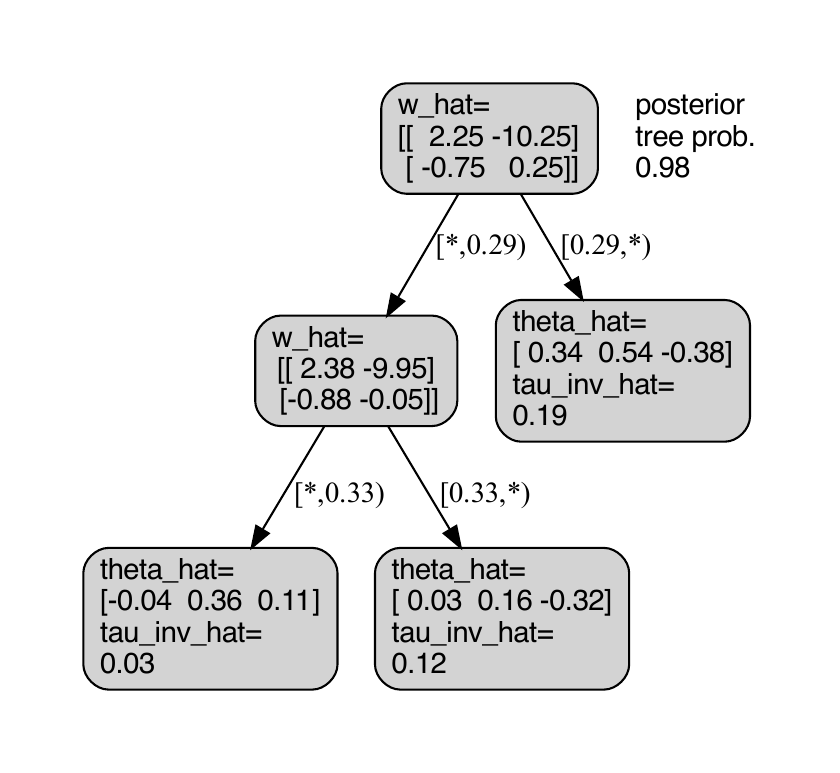}
    \caption{The MAP estimated model and parameters for \texttt{unemp}}
   \label{fig:unemp_tree}
\end{figure}

Figure \ref{fig:unemp_tree} shows the obtained MAP model and parameters. 
The endpoints of the interval written on each edge represent the thresholds obtained by solving $\sigma_1 (\hat{\bm W}_s [1, x]^\top) =\sigma_2 (\hat{\bm W}_s [1, x]^\top)$ for the parent node $s$.
We can see that the same tree as in \cite{PK25} has been estimated, but the thresholds and parameters differ for each node. We used the same initial thresholds as BCT-AR, but they have moved from there to better solutions for the objective function $\mathrm{VL}(q; \bm W) + \ln p(\bm W)$ through the variational inference.

\subsection{Experiment 2: prediction}

Second, we confirm the effect of soft-splitting on prediction accuracy. We use the artificial data \texttt{sim\_1}, \texttt{sim\_2}, \texttt{sim\_3} and real-world data \texttt{unemp}, \texttt{gnp}, \texttt{ibm} employed in \cite{PK25}. 
Moreover, we use our original data \texttt{sim\_1\_soft} and \texttt{sim\_2\_soft}. They are generated from Soft-BCT models where the tree structures and the parameters at the leaf nodes are the same as those of the BCT-AR model used to generate \texttt{sim\_1} and \texttt{sim\_2} in \cite{PK25}, but the splits are soft and their location vary across nodes. For details, please refer to Appendix \ref{appendix:gen_models}. While \cite{PK25} uses only a single dataset in their experiments, we generated these data 30 times from the same models and parameters.
The first $50\%$ of each sequence is used as training data and the rest as test data. 

For Soft-BCT, we use the following hyperparameters: $D_\mathrm{max} = 10$, $g_s=2^{-m}$ for any $s$, $\bm \mu = \bm 0$, $\bm \Lambda = \bm I$, $(a,b)=(1.0, 1.0)$ for \texttt{sim\_2}, \texttt{sim\_3}, and \texttt{gnp}, $(a,b)=(0.1, 0.1)$ for \texttt{sim\_1} and \texttt{unemp}, and $(a,b)=(0.1, 50.0)$ for \texttt{ibm}.\footnote{Although these values are different to those reported in \cite{PK25} and the GitHub repository (see the URL in \cite{PK25}), we confirmed the same MAP models, parameters, and thresholds are obtained from these values by using whole data (including test data). Therefore, there is a possibility that the thresholds reported in \cite{PK25} are estimated by using not only the training data but also the test data.\label{ft:hparams}} 
We set $\bm \eta_m$ in a similar manner to Experiment 1 based on the thresholds $h_1, \dots, h_{M-1}$ obtained from the exhaustive search only on the training data. 
For BCT-AR, we used the same hyperparameters (including thresholds) as Soft-BCT. Note that we do not use the thresholds reported in \cite{PK25} to make a fair comparison.

We first perform learning on the training data. Next, we sequentially repeat prediction and learning on the test data. For prediction, we use the expected value of the approximate posterior predictive distribution. Note that we compute the expected value over all trees. This predicted value $\hat{x}_{t+1}$ can be computed as $\hat{x}_{t+1} = \zeta_{s_\lambda}(\bm x_t^\mathrm{A}, \bm x_t^\mathrm{L})$ using the following recursive function:
\begin{align*}
    &\zeta_s (\bm x_t^\mathrm{A}, \bm x_t^\mathrm{L}) \coloneqq \nonumber \\
    &\begin{cases}
        (1-g'_s) (\bm \mu'_s )^\top \bm x_t^\mathrm{A} \\
        \quad + g'_s \sum_{m=1}^M \sigma_m (\hat{\bm w}_{s,m}^\top \bm x_t^\mathrm{L}) \zeta_{s_m} (\bm x_t^\mathrm{A}, \bm x_t^\mathrm{L}), & s \in \mathcal{I}_\mathrm{max}, \\
        (\bm \mu'_s )^\top \bm x_t^\mathrm{A}, & s \in \mathcal{L}_\mathrm{max}.
    \end{cases}
\end{align*}
In sequential learning, we fix $\bm W$ and iterate updates of $q(\bm U)$ and $q(\bm \theta, \bm \tau, T)$ using only the newly given data point ($x_{t+1}$, $\bm x_t^\mathrm{A}$, $\bm x_t^\mathrm{L}$). 
The hyperparameters of the prior distribution at this time are replaced with the hyperparameters of the approximate posterior distribution at the last time point. Also, the initial values for variational inference are set to the approximate posterior distribution at the last time point. 
For details on such sequential learning, see \cite{StreamingVB}.

Table \ref{tab:forecast_soft} shows the results of this experiment. 
Since each dataset in \cite{PK25} consists of a single dataset, we have not calculated confidence intervals or p-values.
In contrast, since our datasets were each generated 30 times, we have used them to calculate the 95\% confidence intervals for the difference in prediction error and the p-values from paired two-tailed t-tests to determine whether the difference is zero. 
For the datasets in \cite{PK25}, Soft-BCT showed superior performance to BCT-AR only for \texttt{unemp}. However, when data are generated from a model with soft splits whose locations vary across nodes, Soft-BCT shows superior results, as expected.
Therefore, it is likely that the inferior performance of Soft-BCT for the datasets in \cite{PK25} is due to the model being too complex for the data---in other words, overfitting.
However, since Soft-BCT includes BCT-AR as a special case, using BCT-AR as an initial value of the variational inference for Soft-BCT, it should achieve comparable performance given a sufficient sample size, even if the data was suitable for BCT-AR.

\begin{table}[t]
    \centering
    \caption{Results of prediction experiments}
    \label{tab:forecast_soft}
    \begin{tabular}{ccccc}
        \hline
         & MSE of & MSE of & 95\% CI & \\
        Datasets & Soft-BCT & BCT-AR & of diff. & p-value \\
        \hline
        \texttt{sim\_1} & 0.137 & \textbf{0.131} & -- & -- \\
        \texttt{sim\_2} & 0.0507 & \textbf{0.0455} & -- & -- \\
        \texttt{sim\_3} & 1.04 & \textbf{0.998} & -- & -- \\
        \hline
        \texttt{unemp} & \textbf{0.0352} & 0.0367 & -- & -- \\
        \texttt{gnp} & 0.378 & \textbf{0.377} & -- & -- \\
        \texttt{ibm} & 82.4 & \textbf{82.3} & -- & -- \\
        \hline
        \texttt{sim\_1\_soft} & \textbf{0.121} & 0.122 & [0.0001, 0.0011] & 0.0305\\
        \texttt{sim\_2\_soft} & \textbf{0.190} & 0.193 & [0.0004, 0.0062] & 0.0287\\
        \hline
    \end{tabular}
\end{table}

\section{Concluding Remark}
We proposed the soft Bayesian context tree (Soft-BCT) model, which subsumes the BCT-AR model in \cite{PK23}, \cite{PK25} as a special case. We developed the variational inference method for the Soft-BCT and confirmed its effectiveness through numerical experiments. As a data generative model, we considered the autoregressive model as in \eqref{eq_data_generative}. If we change this to another data generative model, we can consider another Soft-BCT model.

\section*{Acknowledgment}
This work was supported by JSPS KAKENHI Grant Numbers JP23K03863, JP25K07732, JP26H02489, JP26K06473, and JP26K17386.

\bibliographystyle{IEEEtran}
\bibliography{refs}

@book{berger1985statistical,
  address = {New York},
  author = {Berger, James O.},
  description = {Amazon.com: Statistical Decision Theory and Bayesian Analysis (Springer Series in Statistics) (9780387960982): James Berger: Books},
  isbn = {0387960988 9780387960982 3540960988 9783540960980},
  publisher = {Springer-Verlag},
  refid = {12053541},
  title = {Statistical decision theory and Bayesian analysis},
  year = 1985
}

@INPROCEEDINGS{NakaharaISIT23,
  author={Nakahara, Yuta and Saito, Shota and Shimada, Koshi and Matsushima, Toshiyasu},
  booktitle={2023 IEEE International Symposium on Information Theory (ISIT)}, 
  title={Hyperparameter Learning of {B}ayesian Context Tree Models}, 
  year={2023},
  volume={},
  number={},
  pages={537-542},
  doi={10.1109/ISIT54713.2023.10206456}}

@Article{full_rooted_trees,
AUTHOR = {Nakahara, Yuta and Saito, Shota and Kamatsuka, Akira and Matsushima, Toshiyasu},
TITLE = {Probability Distribution on Full Rooted Trees},
JOURNAL = {Entropy},
VOLUME = {24},
YEAR = {2022},
NUMBER = {3},
ARTICLE-NUMBER = {328},
ISSN = {1099-4300},
DOI = {10.3390/e24030328}
}

@INPROCEEDINGS{CTW93,
  author={Willems, F.M.J. and Shtarkov, Y.M. and Tjalkens, T.J.},
  booktitle={Proceedings. IEEE International Symposium on Information Theory}, 
  title={Context Tree Weighting : A Sequential Universal Source Coding Procedure for Fsmx Sources}, 
  year={1993},
  volume={},
  number={},
  pages={59-59},
  doi={10.1109/ISIT.1993.748374}}

@ARTICLE{CTW95,
  author={F. M. J. {Willems} and Y. M. {Shtarkov} and T. J. {Tjalkens}},
  journal={IEEE Transactions on Information Theory}, 
  title={The context-tree weighting method: basic properties}, 
  year={1995},
  volume={41},
  number={3},
  pages={653-664},
  doi={10.1109/18.382012}
}

@INPROCEEDINGS{CTW_Matsu_09,
author={T. {Matsushima} and S. {Hirasawa}},
booktitle={2009 IEEE International Symposium on Information Theory},
title={Reducing the space complexity of a {Bayes} coding algorithm using an expanded context tree},
year={2009},
volume={},
number={},
pages={719-723},
doi={10.1109/ISIT.2009.5205677},
ISSN={},
month={June},}

@INPROCEEDINGS{CTW_Matsu_94,
  author={Matsushima, T. and Hirasawa, S.},
  booktitle={Proceedings of 1994 IEEE International Symposium on Information Theory}, 
  title={A {Bayes} coding algorithm using context tree}, 
  year={1994},
  volume={},
  number={},
  pages={386},
  doi={10.1109/ISIT.1994.394633}
}

@INPROCEEDINGS{CTW_Matsu_07,
  author={Matsushima, Toshiyasu and Hirasawa, Shigeich},
  booktitle={2007 IEEE International Symposium on Signal Processing and Information Technology}, 
  title={A Class of Prior Distributions on Context Tree Models and an Efficient Algorithm of the {B}ayes Codes Assuming It}, 
  year={2007},
  volume={},
  number={},
  pages={938-941},
  doi={10.1109/ISSPIT.2007.4458049}}

@ARTICLE{TIT_Matsu_91,
  author={Matsushima, T. and Inazumi, H. and Hirasawa, S.},
  journal={IEEE Transactions on Information Theory}, 
  title={A class of distortionless codes designed by {B}ayes decision theory}, 
  year={1991},
  volume={37},
  number={5},
  pages={1288-1293},
  doi={10.1109/18.133247}}

@INPROCEEDINGS{PK21,
  author={Papageorgiou, I. and Kontoyiannis, I. and Mertzanis, L. and Panotopoulou, A. and Skoularidou, M.},
  booktitle={2021 IEEE International Symposium on Information Theory (ISIT)}, 
  title={Revisiting Context-Tree Weighting for {B}ayesian Inference}, 
  year={2021},
  volume={},
  number={},
  pages={2906-2911},
  doi={10.1109/ISIT45174.2021.9518189}}

@article{kontoyiannis_22,
author = {Kontoyiannis, Ioannis and Mertzanis, Lambros and Panotopoulou, Athina and Papageorgiou, Ioannis and Skoularidou, Maria},
title = {Bayesian context trees: Modelling and exact inference for discrete time series},
journal = {Journal of the Royal Statistical Society: Series B (Statistical Methodology)},
volume = {84},
number = {4},
pages = {1287-1323},
doi = {https://doi.org/10.1111/rssb.12511},
eprint = {https://rss.onlinelibrary.wiley.com/doi/pdf/10.1111/rssb.12511},
year = {2022}
}

@INPROCEEDINGS{PK23,
  author={Papageorgiou, Ioannis and Kontoyiannis, Ioannis},
  booktitle={2023 IEEE International Symposium on Information Theory (ISIT)}, 
  title={Context-tree weighting for real-valued time series: Bayesian inference with hierarchical mixture models}, 
  year={2023},
  volume={},
  number={},
  pages={2464-2469},
  doi={10.1109/ISIT54713.2023.10206739}}

@INPROCEEDINGS{PK_22,
  author={Papageorgiou, Ioannis and Kontoyiannis, Ioannis},
  booktitle={2022 IEEE International Symposium on Information Theory (ISIT)}, 
  title={The Posterior Distribution of {B}ayesian Context-Tree Models: Theory and Applications}, 
  year={2022},
  volume={},
  number={},
  pages={702-707},
  doi={10.1109/ISIT50566.2022.9834791}}

@article{PK25,
title = {The {B}ayesian context trees state space model for time series modelling and forecasting},
journal = {International Journal of Forecasting},
volume = {42},
number = {2},
pages = {474-491},
year = {2026},
issn = {0169-2070},
doi = {https://doi.org/10.1016/j.ijforecast.2025.07.009},
author = {Ioannis Papageorgiou and Ioannis Kontoyiannis},
}

@INPROCEEDINGS{HME,
  author={Jordan, M.I. and Jacobs, R.A.},
  booktitle={Proceedings of 1993 International Conference on Neural Networks (IJCNN-93-Nagoya, Japan)}, 
  title={Hierarchical mixtures of experts and the EM algorithm}, 
  year={1993},
  volume={2},
  number={},
  pages={1339-1344 vol.2},
  doi={10.1109/IJCNN.1993.716791}}

@Book{bishop,
author = {Bishop, Christopher},
title = {Pattern Recognition and Machine Learning},
year = {2006},
month = {January},
publisher = {Springer}
}

@INPROCEEDINGS{CTM,
  author={Volf, P.A.J. and Willems, F.M.J.},
  booktitle={Proceedings of 1995 IEEE International Symposium on Information Theory}, 
  title={On the context tree maximizing algorithm}, 
  year={1995},
  volume={},
  number={},
  pages={20-},
  doi={10.1109/ISIT.1995.531122}}

@INPROCEEDINGS{Gao2006,
  author={Gao, Yun and Kontoyiannis, Ioannis and Bienenstock, Elie},
  booktitle={2006 IEEE International Symposium on Information Theory}, 
  title={From the Entropy to the Statistical Structure of Spike Trains}, 
  year={2006},
  volume={},
  number={},
  pages={645-649},
  doi={10.1109/ISIT.2006.261864}}

@article{CTWclassification,
author = {Begleiter, Ron and El-Yaniv, Ran and Yona, Golan},
title = {On Prediction Using Variable Order Markov Models},
year = {2004},
issue_date = {July 2004},
publisher = {AI Access Foundation},
address = {El Segundo, CA, USA},
volume = {22},
number = {1},
issn = {1076-9757},
journal = {J. Artif. Int. Res.},
month = {dec},
pages = {385–421},
numpages = {37}
}

@Article{CTWentropy,
AUTHOR = {Gao, Yun and Kontoyiannis, Ioannis and Bienenstock, Elie},
TITLE = {Estimating the Entropy of Binary Time Series: Methodology, Some Theory and a Simulation Study},
JOURNAL = {Entropy},
VOLUME = {10},
YEAR = {2008},
NUMBER = {2},
PAGES = {71--99},
ISSN = {1099-4300},
DOI = {10.3390/entropy-e10020071}
}

@INPROCEEDINGS{kontoyiannis_cange_point,
  author={Lungu, Valentinian and Papageorgiou, Ioannis and Kontoyiannis, Ioannis},
  booktitle={2022 IEEE Information Theory Workshop (ITW)}, 
  title={Bayesian Change-Point Detection via Context-Tree Weighting}, 
  year={2022},
  volume={},
  number={},
  pages={125-130},
  doi={10.1109/ITW54588.2022.9965823}}

@INPROCEEDINGS{PK_23_ITW,
  author={Papageorgiou, Ioannis and Kontoyiannis, Ioannis},
  booktitle={2023 IEEE Information Theory Workshop (ITW)}, 
  title={Truly {B}ayesian Entropy Estimation}, 
  year={2023},
  volume={},
  number={},
  pages={497-502},
  doi={10.1109/ITW55543.2023.10161645}}

@article{PK_24,
author = {Ioannis Papageorgiou and Ioannis Kontoyiannis},
title = {{Posterior Representations for Bayesian Context Trees: Sampling, Estimation and Convergence}},
volume = {19},
journal = {Bayesian Analysis},
number = {2},
publisher = {International Society for Bayesian Analysis},
pages = {501 -- 529},
year = {2024},
doi = {10.1214/23-BA1362}
}

@ARTICLE{CTW_98,
  author={Willems, F.M.J.},
  journal={IEEE Transactions on Information Theory}, 
  title={The context-tree weighting method: extensions}, 
  year={1998},
  volume={44},
  number={2},
  pages={792-798},
  doi={10.1109/18.661523}}

@INPROCEEDINGS{CTW_06,
  author={Ignatenko, Tanya and Schrijen, Geert-jan and Skoric, Boris and Tuyls, Pim and Willems, Frans},
  booktitle={2006 IEEE International Symposium on Information Theory}, 
  title={Estimating the Secrecy-Rate of Physical Unclonable Functions with the Context-Tree Weighting Method}, 
  year={2006},
  volume={},
  number={},
  pages={499-503},
  doi={10.1109/ISIT.2006.261765}}

@ARTICLE{K_24,
  author={Kontoyiannis, Ioannis},
  journal={IEEE Transactions on Information Theory}, 
  title={Context-Tree Weighting and {B}ayesian Context Trees: Asymptotic and Non-Asymptotic Justifications}, 
  year={2024},
  volume={70},
  number={2},
  pages={1204-1219},
  doi={10.1109/TIT.2023.3313114}}

@ARTICLE{Goto_98,
  author={Masayuki Gotoh and Toshiyasu Matsushima and Shigeichi Hirasawa },
  journal={IEICE TRANSACTIONS on Fundamentals}, 
  title={A Generalization of {B. S. Clarke} and {A. R. Barron}'s Asymptotics of {B}ayes Codes for FSMX Sources}, 
  year={1998},
  volume={E81-A},
  number={10},
  pages={2123-2132},
  doi={},
  ISSN={},
  month={October},}

@ARTICLE{Goto_01,
  author={Goto, M. and Matsushima, T. and Hirasawa, S.},
  journal={IEEE Transactions on Information Theory}, 
  title={An analysis of the difference of code lengths between two-step codes based on {MDL} principle and {B}ayes codes}, 
  year={2001},
  volume={47},
  number={3},
  pages={927-944},
  doi={10.1109/18.915647}}

@INPROCEEDINGS{Saito_15,
  author={Saito, Shota and Miya, Nozomi and Matsushima, Toshiyasu},
  booktitle={2015 IEEE International Symposium on Information Theory (ISIT)}, 
  title={Fundamental limit and pointwise asymptotics of the {B}ayes code for Markov sources}, 
  year={2015},
  volume={},
  number={},
  pages={1986-1990},
  doi={10.1109/ISIT.2015.7282803}}

@ARTICLE{Saito_15_IEICE,
  author={Shota Saito and Nozomi Miya and Toshiyasu Matsushima },
  journal={IEICE TRANSACTIONS on Fundamentals}, 
  title={Evaluation of the {B}ayes Code from Viewpoints of the Distribution of Its Codeword Lengths}, 
  year={2015},
  volume={E98-A},
  number={12},
  pages={2407-2414},
  doi={10.1587/transfun.E98.A.2407},
  ISSN={1745-1337},
  month={December},}

@ARTICLE{Saito_16,
  author={Shota Saito and Toshiyasu Matsushima },
  journal={IEICE TRANSACTIONS on Fundamentals}, 
  title={Evaluation of Overflow Probability of {B}ayes Code in Moderate Deviation Regime}, 
  year={2017},
  volume={E100-A},
  number={12},
  pages={2728-2731},
  doi={10.1587/transfun.E100.A.2728},
  ISSN={1745-1337},
  month={December},}

@ARTICLE{Miya_14,
  author={Nozomi Miya and Tota Suko and Goki Yasuda and Toshiyasu Matsushima },
  journal={IEICE TRANSACTIONS on Fundamentals}, 
  title={Asymptotics of {B}ayesian Inference for a Class of Probabilistic Models under Misspecification}, 
  year={2014},
  volume={E97-A},
  number={12},
  pages={2352-2360},
  doi={10.1587/transfun.E97.A.2352},
  ISSN={1745-1337},
  month={December},}

@inproceedings{StreamingVB,
 author = {Broderick, Tamara and Boyd, Nicholas and Wibisono, Andre and Wilson, Ashia C and Jordan, Michael I},
 booktitle = {Advances in Neural Information Processing Systems},
 editor = {C.J. Burges and L. Bottou and M. Welling and Z. Ghahramani and K.Q. Weinberger},
 pages = {},
 publisher = {Curran Associates, Inc.},
 title = {Streaming Variational {B}ayes},
 volume = {26},
 year = {2013}
}

@article{ChangingContextTree,
  title={An Efficient {Bayes} Coding Algorithm for Changing Context Tree Model},
  author={Koshi Shimada and Shota Saito and Toshiyasu Matsushima},
  journal={IEICE Transactions on Fundamentals of Electronics, Communications and Computer Sciences},
  volume={E107.A},
  number={3},
  pages={448-457},
  year={2024},
  doi={10.1587/transfun.2023TAP0017}
}

@Article{Quadtree21,
AUTHOR = {Nakahara, Yuta and Matsushima, Toshiyasu},
TITLE = {A Stochastic Model for Block Segmentation of Images Based on the Quadtree and the {Bayes} Code for It},
JOURNAL = {Entropy},
VOLUME = {23},
YEAR = {2021},
NUMBER = {8},
ARTICLE-NUMBER = {991},
PubMedID = {34441131},
ISSN = {1099-4300},
DOI = {10.3390/e23080991}
}

@Article{Quadtree22,
AUTHOR = {Nakahara, Yuta and Matsushima, Toshiyasu},
TITLE = {Stochastic Model of Block Segmentation Based on Improper Quadtree and Optimal Code under the {Bayes} Criterion},
JOURNAL = {Entropy},
VOLUME = {24},
YEAR = {2022},
NUMBER = {8},
ARTICLE-NUMBER = {1152},
PubMedID = {36010816},
ISSN = {1099-4300},
DOI = {10.3390/e24081152}
}

@inproceedings{nakahara_aistats25,
    author = {Nakahara, Yuta and Saito, Shota and Ichijo, Naoki and Kazama, Koki and Matsushima, Toshiyasu},
    booktitle = {Proceedings of The 28th International Conference on Artificial Intelligence and Statistics},
    editor = {Li, Yingzhen and Mandt, Stephan and Agrawal, Shipra and Khan, Emtiyaz},
    month = {03--05 May},
    pages = {1045--1053},
    publisher = {PMLR},
    series = {Proceedings of Machine Learning Research},
    title = {Bayesian Decision Theory on Decision Trees: Uncertainty Evaluation and Interpretability},
    volume = {258},
    year = {2025}}

@article{dobashi_entropy21,
    article-number = {768},
    author = {Dobashi, Nao and Saito, Shota and Nakahara, Yuta and Matsushima, Toshiyasu},
    doi = {10.3390/e23060768},
    issn = {1099-4300},
    journal = {Entropy},
    number = {6},
    pubmedid = {34207209},
    title = {Meta-Tree Random Forest: Probabilistic Data-Generative Model and {Bayes} Optimal Prediction},
    volume = {23},
    year = {2021}}

@inproceedings{ichijo_mlsp25,
    author = {Ichijo, Naoki and Matsushima, Toshiyasu},
    booktitle = {2025 IEEE 35th International Workshop on Machine Learning for Signal Processing (MLSP)},
    doi = {10.1109/MLSP62443.2025.11204340},
    pages = {1-6},
    title = {Meta-Tree: {Bayesian} Approach to Avoid Overfitting in Decision Trees and Analysis on the Application to Boosting},
    year = {2025}}

@ARTICLE{Nakahara_batch_metatree,
  title={Batch Updating of a Posterior Tree Distribution Over a Meta-Tree},
  author={Yuta Nakahara and Toshiyasu Matsushima},
  journal={IEICE Transactions on Fundamentals of Electronics, Communications and Computer Sciences},
  volume={E107.A},
  number={3},
  pages={523-525},
  year={2024},
  doi={10.1587/transfun.2023TAL0003}
}

@article{rooted_trees,
  title={Probability Distribution on Rooted Trees: Generalization from Full Trees},
  author={Yuta Nakahara and Shota Saito and Akira Kamatsuka and Toshiyasu Matsushima},
  journal={IEICE Transactions on Fundamentals of Electronics, Communications and Computer Sciences},
  volume={E109-A},
  number={3},
  pages={},
  year={2025},
  doi={10.1587/transfun.2025TAP0005}
}

@misc{VSBT_arXiv,
author = {Yuta Nakahara and Shota Saito and Kohei Horinouchi and Koshi Shimada and Naoki Ichijo and Manabu Kobayashi and Toshiyasu Matsushima},
title = {Variable Splitting Binary Tree Models based on {Bayesian} Context Tree Models for Time Series Segmentation},
doi={},
year = {arXiv, 2026}
}

\appendices

\section{Lemmas of posterior distributions} \label{section_appendix_posterior}

The following lemmas, Lemma \ref{lem_q_U} and Lemma \ref{lem_q_T_theta_tau}, give the form of the posterior distributions $q(\bm U)$ and $q(T, \bm \theta, \tau)$. Note that the expectation 
\begin{align*}
    \mathbb{E}_{q(T, \bm \theta, \bm \tau)} \left[ I \{ s_m \in \mathcal{L}_T \} \ln \mathcal{N}(x_t | (\bm x_t^\mathrm{A})^\top \bm \theta_{s_m}, \tau_{s_m}^{-1}) \right]
\end{align*}
in Lemma \ref{lem_q_U} and the expectation 
\begin{align*}
    \mathbb{E}_{q(\bm U_t)}[I \{ s \preceq {\sf s}(\bm U_t) \}]
\end{align*}
in Lemma \ref{lem_q_T_theta_tau} are given in Lemma \ref{lem_expectations}.

\begin{Lemma} \label{lem_q_U}
The posterior $q(\bm U)$ can be factorized as
\begin{align*}
    q(\bm U) = \prod_{t=1}^{n} q(\bm U_t)
\end{align*}
and each $q^*(\bm U_t)$ is 
    \begin{align}
        q(\bm U_t) = \prod_{s \in \mathcal{I}_\mathrm{max}} \prod_{m=1}^M (\pi'_{t,s,s_m})^{I \{ s_m \preceq {\sf s}(\bm U_t) \}}, \label{eq_update_formula_q_star}
    \end{align}
where 
\begin{align}
    \pi'_{t, s, s_m} \coloneqq \frac{\rho_{t, s, s_m}}{\sum_{m=1}^M \rho_{t, s, s_m}} \label{lemma_pi}
\end{align}
and 
\begin{align}
    &\ln \rho_{t, s, s_m} \coloneqq \nonumber \\
    &\begin{cases}
        \ln \sigma_m (\bm W_s \bm x_t^\mathrm{L}) \\
        +\mathbb{E}_{q(T, \bm \theta, \bm \tau)} \left[ I \{ s \in \mathcal{L}_T \} \ln \mathcal{N}(x_t | (\bm x_t^\mathrm{A})^\top \bm \theta_s, \tau_s^{-1}) \right] \\ 
        + \ln \sum_{s_\mathrm{ch} \in \mathrm{Ch}(s_m)} \rho_{t,s_m,s_\mathrm{ch}}, \\  \qquad \qquad \qquad \qquad \qquad \qquad \qquad (\mathrm{if} ~ s_m \in \mathcal{I}_\mathrm{max}), \\
        \\
        \ln \sigma_m (\bm W_s \bm x_t^\mathrm{L}) \\
        + \mathbb{E}_{q(T, \bm \theta, \bm \tau)} \left[ I \{ s \in \mathcal{L}_T \} \ln \mathcal{N}(x_t | (\bm x_t^\mathrm{A})^\top \bm \theta_s, \tau_s^{-1}) \right], \\  
        \qquad \qquad \qquad \qquad \qquad \qquad \qquad (\mathrm{if} ~ s_m \in \mathcal{L}_\mathrm{max}).
    \end{cases} \label{lemma_rho}
\end{align} 
\end{Lemma}

\begin{IEEEproof}
In this proof, ``$\mathrm{const.}$'' denotes constants that do not depend on $\bm U = \{\bm U_t \}_{t=1}^n$. For the sake of simplicity, we use $\mathbb{E}_{q}$ instead of $\mathbb{E}_{q(T, \bm \theta, \bm \tau)}$. 

First, from \eqref{eq_q_star_u}, we have
\begin{align*}
    &\ln q(\bm U) \\
    & \quad = \mathbb{E}_{q} \bigl[\ln p(\bm x, \bm U, T, \bm \theta, \bm \tau | \bm W) \bigr] + \mathrm{const.} \\
    & \quad = \mathbb{E}_{q} \bigl[\ln p(T) p(\bm \theta, \bm \tau | T) p(\bm x, \bm U | \bm W, T, \bm \theta, \bm \tau) \bigr] + \mathrm{const.} \\
    & \quad = \mathbb{E}_{q} \bigl[\ln p(\bm x, \bm U | \bm W, T, \bm \theta, \bm \tau) \bigr] + \mathrm{const.}
\end{align*}
By substituting \eqref{eq_context_model} and \eqref{eq_data_generative}, we see that $q(\bm U)$ has the form 
\begin{align*}
q(\bm U) = \prod_{t=1}^{n} q(\bm U_t), 
\end{align*}
and each $q(\bm U_t)$ is written as
\begin{align}
    &\ln q(\bm U_t) \nonumber \\
    &= \sum_{s \in \mathcal{I}_\mathrm{max}} \sum_{m=1}^M I \{ s_m \preceq {\sf s}(\bm U_t) \} \ln \sigma_m (\bm W_s \bm x_t^\mathrm{L}) \nonumber \\
    &\quad + \sum_{s \in \mathcal{S}_\mathrm{max}} I \{ s \preceq {\sf s}(\bm U_t) \} \nonumber \\
    & \qquad \times \mathbb{E}_q \left[ I \{ s \in \mathcal{L}_T \} \ln \mathcal{N}(x_t | (\bm x_t^\mathrm{A})^\top \bm \theta_s, \tau_s^{-1}) \right] \nonumber \\
    & \qquad + \mathrm{const.} \nonumber \\
    &= \sum_{s \in \mathcal{I}_\mathrm{max}} \sum_{m=1}^M I \{ s_m \preceq {\sf s}(\bm U_t) \} \ln \sigma_m (\bm W_s \bm x_t^\mathrm{L}) \nonumber \\
    &\quad + \mathbb{E}_q \left[ I \{ s_\lambda \in \mathcal{L}_T \} \ln \mathcal{N}(x_t | (\bm x_t^\mathrm{A})^\top \bm \theta_s, \tau_s^{-1}) \right] \nonumber \\
    &\quad + \sum_{s \in \mathcal{I}_\mathrm{max}} \sum_{m=1}^M I \{ s_m \preceq {\sf s}(\bm U_t) \} \nonumber \\
    & \qquad \times \mathbb{E}_q \left[ I \{ s_m \in \mathcal{L}_T \} \ln \mathcal{N}(x_t | (\bm x_t^\mathrm{A})^\top \bm \theta_{s_m}, \tau_{s_m}^{-1}) \right] \nonumber \\
    &\quad + \mathrm{const.} \nonumber \\
    &= \sum_{s \in \mathcal{I}_\mathrm{max}} \sum_{m=1}^M I \{ s_m \preceq {\sf s}(\bm U_t) \} \Biggl\{ \ln \sigma_m (\bm W_s \bm x_t^\mathrm{L}) \nonumber \\
    & \quad + \mathbb{E}_q \left[ I \{ s_m \in \mathcal{L}_T \} \ln \mathcal{N}(x_t | (\bm x_t^\mathrm{A})^\top \bm \theta_{s_m}, \tau_{s_m}^{-1}) \right] \Biggr\} \nonumber \\
    & \quad + \mathrm{const.} \label{eq_q_u_goal}
\end{align}

Next, we calculate the logarithm of the right-hand side of \eqref{eq_update_formula_q_star}:
\begin{align}
    \sum_{s \in \mathcal{I}_\mathrm{max}} \sum_{m=1}^M I \{ s_m \preceq {\sf s}(\bm U_t) \}  \ln \pi'_{t,s,s_m}. \label{eq_q_u_calc}
\end{align}
Our goal is to show that \eqref{eq_q_u_calc} is the same as \eqref{eq_q_u_goal}. In the following, $\mathcal{S}_d$ denotes the set of nodes whose depth is $d$, and $\mathcal{S}_{<d}$ denotes the set of nodes whose depth is less than $d$. 

By substituting \eqref{lemma_pi} and \eqref{lemma_rho} into \eqref{eq_q_u_calc}, we have
\begin{align}
    &\sum_{s \in \mathcal{I}_\mathrm{max}} \sum_{m=1}^M I \{ s_m \preceq {\sf s}(\bm U_t) \}  \ln \pi'_{t,s,s_m} \nonumber \\
    &= \sum_{s \in \mathcal{S}_{<D_\mathrm{max}-1}} \sum_{m=1}^M I \{ s_m \preceq {\sf s}(\bm U_t) \}  \ln \pi'_{t,s,s_m} \nonumber \\
    &\quad + \sum_{s \in \mathcal{S}_{D_\mathrm{max}-1}} \sum_{m=1}^M I \{ s_m \preceq {\sf s}(\bm U_t) \}  \ln \pi'_{t,s,s_m} \nonumber \\
    &= \sum_{s \in \mathcal{S}_{<D_\mathrm{max}-1}} \sum_{m=1}^M I \{ s_m \preceq {\sf s}(\bm U_t) \} \Biggl\{ \ln \sigma_m (\bm W_s \bm x_t^\mathrm{L}) \nonumber \\
    &\qquad  +\mathbb{E}_q \left[ I \{ s \in \mathcal{L}_T \} \ln \mathcal{N}(x_t | (\bm x_t^\mathrm{A})^\top \bm \theta_s, \tau_s^{-1}) \right] \nonumber \\
    & \qquad + \ln \sum_{s_\mathrm{ch} \in \mathrm{Ch}(s_m)} \rho_{t,s_m,s_\mathrm{ch}} - \ln \sum_{m=1}^M \rho_{t, s, s_m} \Biggr\} \nonumber \\
    &\quad + \sum_{s \in \mathcal{S}_{D_\mathrm{max}-1}} \sum_{m=1}^M I \{ s_m \preceq {\sf s}(\bm U_t) \}  \Biggl\{ \ln \sigma_m (\bm W_s \bm x_t^\mathrm{L}) \nonumber \\
    &\qquad  +\mathbb{E}_q \left[ I \{ s \in \mathcal{L}_T \} \ln \mathcal{N}(x_t | (\bm x_t^\mathrm{A})^\top \bm \theta_s, \tau_s^{-1}) \right] \nonumber \\
    &\qquad - \ln \sum_{m=1}^M \rho_{t, s, s_m} \Biggr\} \nonumber \\
    &= \sum_{s \in \mathcal{I}_\mathrm{max}} \sum_{m=1}^M I \{ s_m \preceq {\sf s}(\bm U_t) \}  \Biggl\{ \ln \sigma_m (\bm W_s \bm x_t^\mathrm{L}) \nonumber \\
    &\quad + \mathbb{E}_q \left[ I \{ s \in \mathcal{L}_T \} \ln \mathcal{N}(x_t | (\bm x_t^\mathrm{A})^\top \bm \theta_s, \tau_s^{-1}) \right] \Biggr\} \nonumber \\
    &\quad - \sum_{m=1}^M I \{ (s_\lambda)_m \preceq {\sf s}(\bm U_t) \} \ln \sum_{m'=1}^M \rho_{t, s_\lambda, (s_\lambda)_{m'}}, \label{eq_q_u_with_const}
\end{align}
where the last equality follows because $\ln \sum_{m=1}^M \rho_{t, s, s_m}$ are canceled for $s$ except $s_\lambda$. Moreover, since
\begin{align*}
    &\sum_{m=1}^M I \{ (s_\lambda)_m \preceq {\sf s}(\bm U_t) \} \ln \sum_{m'=1}^M \rho_{t, s_\lambda, (s_\lambda)_{m'}} \\
    & \quad = \ln \sum_{m'=1}^M \rho_{t, s_\lambda, (s_\lambda)_{m'}},
\end{align*}
the last term of \eqref{eq_q_u_with_const} is a constant that does not depend on $\bm U_t$. Therefore, we conclude that \eqref{eq_q_u_goal} and \eqref{eq_q_u_with_const} are the same.
\end{IEEEproof}

\begin{Lemma} \label{lem_q_T_theta_tau}
Let
\begin{align}
    &\bm Q_s \coloneqq\mathrm{diag} \Big\{ \mathbb{E}_{q(\bm U_1)}[I \{ s \preceq {\sf s}(\bm U_1) \}], \mathbb{E}_{q(\bm U_2)}[I \{ s \preceq {\sf s}(\bm U_2) \}], \nonumber \\
    & \qquad \qquad \qquad \qquad \qquad \qquad \cdots, \mathbb{E}_{q(\bm U_n)}[I \{ s \preceq {\sf s}(\bm U_n) \}] \Big \}. \label{lemma_Q_s}
\end{align}
The posterior $q(T, \bm \theta, \tau)$ can be factorized as $q(T, \bm \theta, \bm \tau) = q(T) \prod_{s \in \mathcal{L}_T} q(\bm \theta_s, \tau_s)$. For each $s \in \mathcal{L}_T$, $q(\bm \theta_s, \tau_s)$ is 
\begin{align*}
    q(\bm \theta_s, \tau_s) = \mathcal{N}(\bm \theta_s | \bm \mu'_s, (\tau_s \bm \Lambda'_s)^{-1}) \mathrm{Gam}(\tau_s | a'_s, b'_s),
\end{align*}
where 
\begin{align}
        \bm \Lambda'_s &\coloneqq \bm \Lambda + (\bm X^\mathrm{A})^\top \bm Q_s \bm X^\mathrm{A}, \label{lemma_Lambda}\\
        \bm \mu'_s &\coloneqq \left( \bm \Lambda'_s \right)^{-1} \left( \bm \Lambda \bm \mu + (\bm X^\mathrm{A})^\top \bm Q_s \bm x \right), \label{lemma_mu}\\
        a'_s &\coloneqq a + \frac{1}{2}\mathrm{Tr} \{ \bm Q_s \}, \label{lemma_a}\\
        b'_s &\coloneqq b + \frac{1}{2} \left( \bm \mu^\top \bm \Lambda \bm \mu + \bm x^\top \bm Q_s \bm x - (\bm \mu'_s)^\top \bm \Lambda'_s \bm \mu'_s \right), \label{lemma_b}
\end{align}
and $q(T)$ is 
\begin{align*}
    q(T) = \left( \prod_{s \in \mathcal{I}_T} g'_s \right) \left( \prod_{s \in \mathcal{L}_T} (1-g'_s) \right), 
\end{align*}
where 
\begin{align}
    g'_s &\coloneqq \begin{cases}
    \frac{g_s \prod_{s_\mathrm{ch} \in \mathrm{Ch}(s)} \phi_{s_\mathrm{ch}}}{\phi_s}, & s \in \mathcal{I}_\mathrm{max}, \\
    0, & s \in \mathcal{L}_\mathrm{max},
    \end{cases} \label{lemma_g}\\
    \phi_s &\coloneqq \begin{cases}
    (1-g_s) \gamma_s + g_s \prod_{s_\mathrm{ch} \in \mathrm{Ch}(s)} \phi_{s_\mathrm{ch}}, & s \in \mathcal{I}_\mathrm{max}, \\
    \gamma_s, & s \in \mathcal{L}_\mathrm{max}.
    \end{cases} \nonumber \\
    \ln \gamma_s &= \frac{1}{2} \ln |\bm \Lambda| - \frac{1}{2} \ln |\bm \Lambda'_s| + a \ln b - a'_s \ln b'_s \nonumber \\
    & \quad - \ln \Gamma (a) + \ln \Gamma (a'_s) - \frac{1}{2} \mathrm{Tr} \{ \bm Q_s \} \ln 2\pi. \label{lemma_gamma}
\end{align}
\end{Lemma}

\begin{IEEEproof}
In this proof, ``$\mathrm{const.}$'' denotes constants that do not depend on $T$, $\bm \theta$, and $\bm \tau$. First, from \eqref{eq_q_star_T_theta_tau}, we have 
\begin{align}
    &\ln q(T, \bm \theta, \bm \tau) \nonumber \\
    &= \ln p(T) + \ln p(\bm \theta, \bm \tau | T) + \mathbb{E}_{q(\bm U)}[ \ln p(\bm x | \bm U, \bm \theta, \bm \tau, T)] \nonumber \\
    & \quad + \mathrm{const.} \nonumber \\
    &= \sum_{s \in \mathcal{I}_\mathrm{T}} \ln g_s + \sum_{s \in \mathcal{L}_T} \ln (1-g_s) + \sum_{s \in \mathcal{L}_T} \ln \gamma_s \nonumber \\
    & \quad + \sum_{s \in \mathcal{L}_T} \ln \mathcal{N}(\bm \theta_s | \bm \mu'_s, (\tau_s \bm \Lambda'_s)^{-1})\mathrm{Gam}(\tau_s | a'_s, b'_s) + \mathrm{const.}, \label{q_T_theta_tau_form}
\end{align}
where $\bm \Lambda'_s$, $\bm \mu'_s$, $a'_s$, $b'_s$, and $\gamma_s$ are defined as in \eqref{lemma_Lambda}, \eqref{lemma_mu}, \eqref{lemma_a}, \eqref{lemma_b}, and \eqref{lemma_gamma}. 

From \eqref{q_T_theta_tau_form}, we see that $q(T, \bm \theta, \bm \tau)$ can be factorized as
\begin{align*}
    q(T, \bm \theta, \bm \tau) = q(T) \prod_{s \in \mathcal{L}_T} q(\bm \theta_s, \tau_s),
\end{align*}
and $q(\bm \theta_s, \tau_s)$ is given as
\begin{align}
    q(\bm \theta_s, \tau_s) = \mathcal{N}(\bm \theta_s | \bm \mu'_s, (\tau_s \bm \Lambda'_s)^{-1})\mathrm{Gam}(\tau_s | a'_s, b'_s).
\end{align}

Furthermore, $q(T)$ corresponds to the posterior distribution when considering $\left(\prod_{s \in \mathcal{I}_T} g_s \right) \left(\prod_{s \in \mathcal{L}_T}(1-g_s) \right)$ as the prior distribution and $\prod_{s \in \mathcal{L}_T} \gamma_s$ as the likelihood function. From \cite[Theorem 7]{full_rooted_trees}, such posterior distribution can be represented as
\begin{align*}
    q(T) = \left( \prod_{s \in \mathcal{I}_T} g'_s \right) \left( \prod_{s \in \mathcal{L}_T} (1-g'_s) \right),
\end{align*}
where $g'_s$ is defined as in \eqref{lemma_g}.
\end{IEEEproof}

\section{Lemma on the calculation of the expectations} \label{section_appendix_expectation}

The expectations in Lemma \ref{lem_q_U} and Lemma \ref{lem_q_T_theta_tau} are calculated as follows.

\begin{Lemma} \label{lem_expectations}
The expectation in \eqref{lemma_rho} is given as
    \begin{align}
        &\mathbb{E}_{q(T, \bm \theta, \bm \tau)} \left[ I \{ s_m \in \mathcal{L}_T \} \ln \mathcal{N}(x_t | (\bm x_t^\mathrm{A})^\top \bm \theta_{s_m}, \tau_{s_m}^{-1}) \right] \nonumber \\
        &=\frac{1}{2} (1-g'_{s_m}) \left(\prod_{\tilde{s} \prec s_m} g'_{\tilde{s}} \right) \Big \{ (-\ln 2\pi + \psi(a'_{s_m}) - \ln b'_{s_m}) \nonumber \\
        & - \frac{a'_{s_m}}{b'_{s_m}}(x_t - (\bm x_t^\mathrm{A})^\top \bm \mu'_{s_m})^2 - (\bm x_t^\mathrm{A})^\top (\bm \Lambda'_{s_m})^{-1} (\bm x_t^\mathrm{A}) \Big \}. \label{eq_expectation_lemma_1}
    \end{align}
Also, the expectation in \eqref{lemma_Q_s} is given as
\begin{align}
    \mathbb{E}_{q(\bm U_t)}[I \{ s \preceq {\sf s}(\bm U_t) \}] =\prod_{s', s'_\mathrm{ch} \preceq s} \pi'_{t, s', s'_\mathrm{ch}}. \label{eq_expectation_lemma_2}
\end{align}
\end{Lemma}

\begin{IEEEproof}
First, we show \eqref{eq_expectation_lemma_1}. Since $q(T, \bm \theta, \bm \tau)=q(T)q(\bm \theta, \bm \tau)$, we have
\begin{align}
    & \mathbb{E}_{q(T, \bm \theta, \bm \tau)} \left[ I \{ s_m \in \mathcal{L}_T \} \ln \mathcal{N}(x_t | (\bm x_t^\mathrm{A})^\top \bm \theta_{s_m}, \tau_{s_m}^{-1}) \right] \nonumber \\
    &\quad = \mathbb{E}_{q(T)} \left[ I \{ s_m \in \mathcal{L}_T \} \right] \mathbb{E}_{q(\bm \theta, \bm \tau)} \left[ \ln \mathcal{N}(x_t | (\bm x_t^\mathrm{A})^\top \bm \theta_{s_m}, \tau_{s_m}^{-1}) \right]. \label{eq_calc_expectation}
\end{align}

From \cite[Theorem 2]{full_rooted_trees}, the expectation $\mathbb{E}_{q(T)} \left[ I \{ s_m \in \mathcal{L}_T \} \right]$ in \eqref{eq_calc_expectation} is given as
\begin{align}
    \mathbb{E}_{q(T)} \left[ I \{ s_m \in \mathcal{L}_T \} \right] = (1-g'_{s_m}) \left(\prod_{\tilde{s} \prec s_m} g'_{\tilde{s}} \right). \label{eq_calc_expectation_q(T)}
\end{align}

The expectation $\mathbb{E}_{q(\bm \theta, \bm \tau)} \left[ \ln \mathcal{N}(x_t | (\bm x_t^\mathrm{A})^\top \bm \theta_{s_m}, \tau_{s_m}^{-1}) \right]$ in \eqref{eq_calc_expectation} is calculated as
\begin{align}
    & \mathbb{E}_{q(\bm \theta, \bm \tau)} \left[ \ln \mathcal{N}(x_t | (\bm x_t^\mathrm{A})^\top \bm \theta_{s_m}, \tau_{s_m}^{-1}) \right] \nonumber \\
    &\quad = -\frac{1}{2} \ln (2 \pi) + \frac{1}{2} \mathbb{E}_{q(\bm \tau)} [\ln (\tau_{s_m})] \nonumber \\
    & \qquad \qquad - \mathbb{E}_{q(\bm \theta, \bm \tau)} \left[\frac{\tau_{s_m}}{2}(x_t - (\bm x_t^\mathrm{A})^\top \bm \theta_{s_m})^2 \right] \nonumber \\
    &\quad =  -\frac{1}{2} \ln (2 \pi) + \frac{1}{2} (\psi(a'_{s_m}) - \ln b'_{s_m}) \nonumber \\
    & \qquad \qquad -\frac{1}{2} \mathbb{E}_{q(\bm \theta, \bm \tau)} \left[ \tau_{s_m} \Big \{ x_t^2 - 2 x_t (\bm x_t^\mathrm{A})^\top \bm \theta_{s_m} \right. \nonumber \\
    & \qquad \qquad \qquad \qquad \qquad \left. + (\bm x_t^\mathrm{A})^\top \bm \theta_{s_m} \bm \theta_{s_m}^\top (\bm x_t^\mathrm{A}) \Big \} \right], \label{eq_calc_expectation_lnNormal}
\end{align}
and the last term in \eqref{eq_calc_expectation_lnNormal} is
\begin{align}
    & \mathbb{E}_{q(\bm \theta, \bm \tau)} \left[ \tau_{s_m} \Big \{ x_t^2 - 2 x_t (\bm x_t^\mathrm{A})^\top \bm \theta_{s_m}+ (\bm x_t^\mathrm{A})^\top \bm \theta_{s_m} \bm \theta_{s_m}^\top (\bm x_t^\mathrm{A}) \Big \} \right] \nonumber \\
    & = \mathbb{E}_{q(\bm \tau)} \Big[\tau_{s_m} \nonumber \\
    & \quad \times \mathbb{E}_{q(\bm \theta |\bm \tau)} \Big[x_t^2 - 2 x_t (\bm x_t^\mathrm{A})^\top \bm \theta_{s_m}+ (\bm x_t^\mathrm{A})^\top \bm \theta_{s_m} \bm \theta_{s_m}^\top (\bm x_t^\mathrm{A}) \Big] \Big] \nonumber \\
    & = \mathbb{E}_{q(\bm \tau)} \Big[\tau_{s_m} \Big( x_t^2 - 2 x_t (\bm x_t^\mathrm{A})^\top \mathbb{E}_{q(\bm \theta |\bm \tau)} [\bm \theta_{s_m}] \nonumber \\
    & \qquad \qquad \qquad+ (\bm x_t^\mathrm{A})^\top \mathbb{E}_{q(\bm \theta |\bm \tau)} [\bm \theta_{s_m} \bm \theta_{s_m}^\top] (\bm x_t^\mathrm{A}) \Big) \Big] \nonumber \\
    &=  \mathbb{E}_{q(\bm \tau)} \Big[\tau_{s_m} \Big( x_t^2 - 2 x_t (\bm x_t^\mathrm{A})^\top \bm \mu'_{s_m} \nonumber \\
    & \qquad \qquad + (\bm x_t^\mathrm{A})^\top \left \{\frac{1}{\tau_{s_m} }(\bm \Lambda'_{s_m})^{-1} + \bm \mu'_{s_m} (\bm \mu'_{s_m})^\top \right \}(\bm x_t^\mathrm{A}) \Big) \Big] \nonumber \\
    &=  \mathbb{E}_{q(\bm \tau)} \Big[\tau_{s_m} \Big( x_t^2 - 2 x_t (\bm x_t^\mathrm{A})^\top \bm \mu'_{s_m} + ((\bm x_t^\mathrm{A})^\top  \bm \mu'_{s_m})^2 \Big) \Big] \nonumber \\
    & \qquad + (\bm x_t^\mathrm{A})^\top (\bm \Lambda'_{s_m})^{-1}(\bm x_t^\mathrm{A}) \nonumber \\
    &=  \mathbb{E}_{q(\bm \tau)} \Big[\tau_{s_m}(x_t - (\bm x_t^\mathrm{A})^\top \bm \mu'_{s_m})^2 \Big] + (\bm x_t^\mathrm{A})^\top (\bm \Lambda'_{s_m})^{-1}(\bm x_t^\mathrm{A}) \nonumber \\
    &=\frac{a'_{s_m}}{b'_{s_m}}(x_t - (\bm x_t^\mathrm{A})^\top \bm \mu'_{s_m})^2+ (\bm x_t^\mathrm{A})^\top (\bm \Lambda'_{s_m})^{-1}(\bm x_t^\mathrm{A}). \label{eq_calc_expectation_lnNormal_2}
\end{align}

Combining \eqref{eq_calc_expectation}, \eqref{eq_calc_expectation_q(T)}, \eqref{eq_calc_expectation_lnNormal}, and \eqref{eq_calc_expectation_lnNormal_2}, we have \eqref{eq_expectation_lemma_1}.

Also, the direct computation of $\mathbb{E}_{q^*(\bm U_t)}[I \{ s \preceq {\sf s}(\bm U_t) \}]$ yields \eqref{eq_expectation_lemma_2}.
\end{IEEEproof}

\section{Details of update formula \texorpdfstring{\eqref{eq_update_formula_w}}{(18)}} \label{section_appendix_w}

The term $\mathrm{VL}(q; \bm W) + \ln p(\bm W)$ in \eqref{eq_estimator_W} is written as
\begin{align*}
    & \mathrm{VL}(q; \bm W) + \ln p(\bm W) \\
    & \quad = \mathbb{E}_{q(\bm U, T, \bm \theta, \bm \tau)}[\ln p(\bm U | \bm W, \bm x)] + \ln p(\bm W) + \mathrm{const.},
\end{align*}
where $\mathrm{const.}$ denotes the terms that do not depend on $\bm W$.

Eliminating terms independent of $\bm W$, such as the normalization constant, yields the following expression of $\mathrm{VL}(q; \bm W) + \ln p(\bm W)$:
\begin{align}
    & \sum_{t=1}^n \sum_{s \in \mathcal{I}_\mathrm{max}} q_{s,t} \sum_{m=1}^M \pi'_{t, s, s_m} \ln \sigma_m (\bm W_s \bm x_t^\mathrm{L}) \nonumber \\
    & \quad - \frac{1}{2} \sum_{s \in \mathcal{I}_\mathrm{max} }\sum_{m=1}^M (\bm w_{s,m} - \bm \eta_m)^\top \bm L (\bm w_{s,m} - \bm \eta_m). \label{eq_VL+lnp}
\end{align}

For each $s$, \eqref{eq_VL+lnp} is the same form of the objective function of usual multiclass logistic regression with a regularization term. Hence, we can apply the same learning algorithm as that of multiclass logistic regression (see \cite[Section 4.3.4]{bishop}). The specific update formula can be derived as follows.

First, we write \eqref{eq_VL+lnp} as
\begin{align*}
    \sum_{s \in \mathcal{I}_\mathrm{max}} (- E(\bm w_s)),
\end{align*}
where $E(\bm w_s)$ is defined as in \eqref{eq_def_E}. Therefore, maximizing $\mathrm{VL}(q; \bm W) + \ln p(\bm W)$ with respect to $\bm w_{s,m}$ is equivalent to minimizing $E(\bm w_s)$ with respect to $\bm w_{s,m}$. 

By differentiating $E(\bm w_s)$ with respect to $\bm w_{s,m}$, we have
\begin{align*}
    & \nabla_{\bm w_{s,m}}E(\bm w_s) =  \\
    & \quad -\sum_{t=1}^n q_{s,t} (\pi'_{t, s, s_m} - \sigma_m (\bm W_s \bm x_t^\mathrm{L})) \bm x_t^\mathrm{L} + \bm L (\bm w_{s,m} - \bm \eta_m).
\end{align*}
The gradient $\nabla E(\bm w_s)$ of $E(\bm w_s)$ with respect to all $\bm w_s$ is a $(J+1)M$-dimensional vector obtained by arranging these for $m=1, \dots , M$.

Next, differentiating $\nabla_{\bm w_{s,m}}E(\bm w_s)$ with respect to $\bm w_{s,m'}$ yields the following $(J+1) \times (J+1)$ matrix.
\begin{align}
    & \nabla_{\bm w_{s,m'}} \nabla_{\bm w_{s,m}} E(\bm w_s) = \sum_{t=1}^n q_{s,t}  \sigma_{m'}(\bm W_s \bm x_t^\mathrm{L}) \nonumber \\
    & \quad \times (\delta_{m,m'} - \sigma_m (\bm W_s \bm x_t^\mathrm{L})) \bm x_t^\mathrm{L} (\bm x_t^\mathrm{L})^\top + \delta_{m,m'} \bm L, \label{eq_Hessian}
\end{align}
where 
\begin{align*}
    \delta_{m,m'} \coloneqq 
    \begin{cases}
        1, & m=m', \\
        0, & m \neq m'.
    \end{cases}
\end{align*}
The Hessian matrix $\bm H_{\bm w_s}$ of $E(\bm w_s)$ is a $(J+1)M \times (J+1)M$ matrix arranged \eqref{eq_Hessian}.

By using $\nabla E(\bm w_s)$ and $\bm H_{\bm w_s}$, we obtain the Newton-Raphson update formula \eqref{eq_update_formula_w} (see also Eq. (4.92) in \cite{bishop}).

\section{The Soft-BCT for generating the data used in Experiment 2}\label{appendix:gen_models}

\begin{figure}
    \centering
    \includegraphics[width=0.5\linewidth]{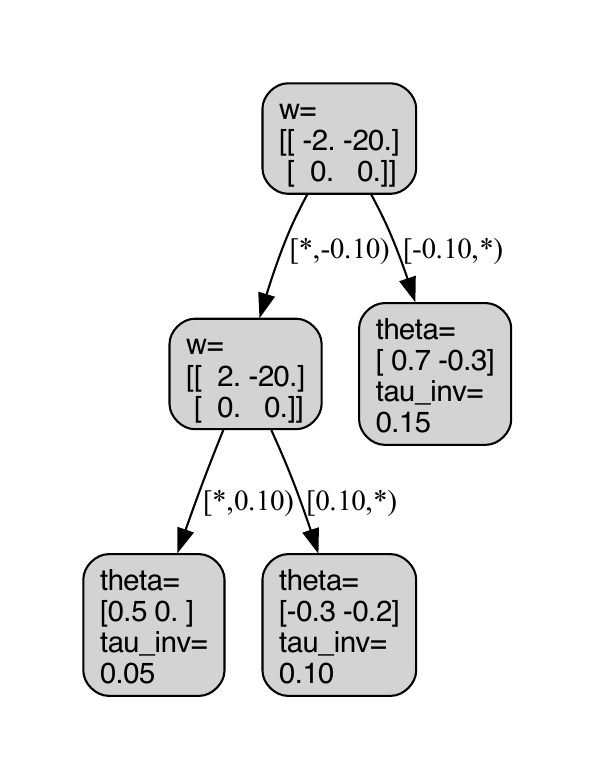}
    \caption{The Soft-BCT used to generate \texttt{sim\_1\_soft} in Experiment 2.}
    \label{fig:sim_1_soft}
\end{figure}

\begin{figure*}
    \centering
    \includegraphics[width=0.8\linewidth]{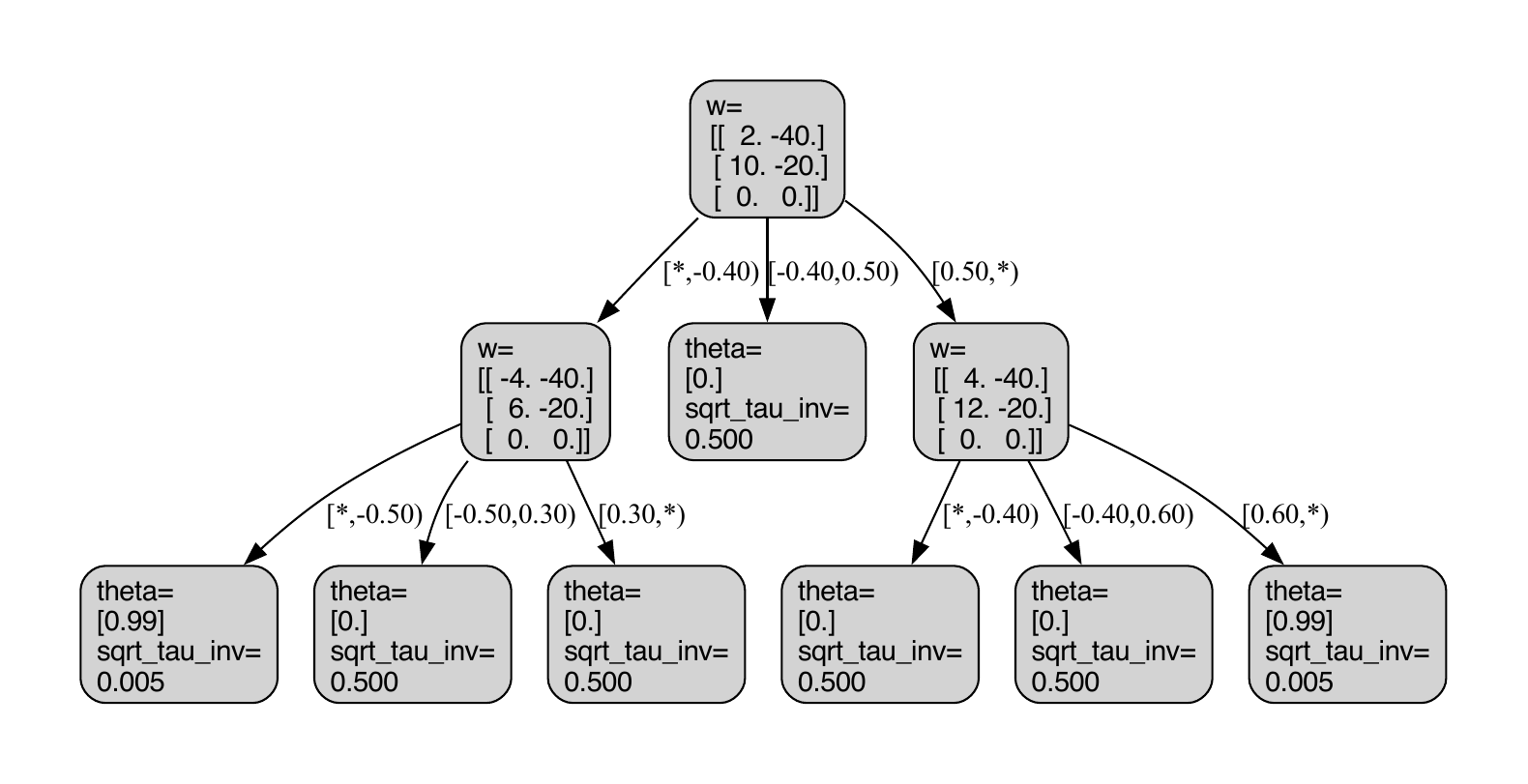}
    \caption{The Soft-BCT used to generate \texttt{sim\_2\_soft} in Experiment 2.}
    \label{fig:sim_2_soft}
\end{figure*}

Figures \ref{fig:sim_1_soft} and \ref{fig:sim_2_soft} show the Soft-BCTs used to generate \texttt{sim\_1\_soft} and \texttt{sim\_2\_soft}, respectively. An interval written on the $m$-th edge of a node $s$ represents the interval of $x$ where $\sigma_m (\bm W_s [1, x]^\top) \geq \sigma_{m'} (\bm W_s [1, x]^\top)$ holds for any $m' \neq m$. The tree structures and the parameters at the leaf nodes are the same as those of the BCT-AR model used to generate \texttt{sim\_1} and \texttt{sim\_2} in \cite{PK25}, but the splits are soft and their location vary across nodes.

\end{document}